\documentclass{article}

\newif\ifcorlcameraready

\usepackage{xcolor} 

\ifcorlcameraready
  \usepackage[final]{corl_2026}
\else
  \usepackage[preprint]{corl_2026}
\fi

\usepackage{booktabs}
\usepackage{makecell}
\usepackage{amsmath}
\usepackage{amssymb}
\usepackage{float}
\usepackage{graphicx}
\usepackage{multirow} 
\usepackage{caption}
\usepackage{wrapfig}
\usepackage{enumitem}

\usepackage{microtype}

\ifcorlcameraready\else
  \newcommand{\arxivdisplayspacing}{%
    \setlength{\abovedisplayskip}{5pt plus 1pt minus 1pt}%
    \setlength{\belowdisplayskip}{5pt plus 1pt minus 1pt}%
    \setlength{\abovedisplayshortskip}{0pt plus 1pt}%
    \setlength{\belowdisplayshortskip}{3pt plus 1pt minus 1pt}%
  }
  \AddToHook{cmd/normalsize/after}{\arxivdisplayspacing}
  \AddToHook{cmd/small/after}{\arxivdisplayspacing}
  \AtBeginDocument{\arxivdisplayspacing}
\fi

\newcommand{\rewardmethod}{WARP}
\newcommand{\progressmodel}{WARP-RM} 
\newcommand{\bcmethod}{WARP-BC}

\newif\ifshowdraftnotes
\showdraftnotesfalse  

\ifshowdraftnotes
    \newcommand{\karim}[1]{\textcolor{blue}{\textbf{[KE: #1]}}}
    \newcommand{\kavish}[1]{\textcolor{brown}{\textbf{[KK: #1]}}}
    \newcommand{\andrew}[1]{\textcolor{red}{\textbf{[AG: #1]}}}
    \newcommand{\jy}[1]{\textcolor{orange}{\textbf{[JY: #1]}}}
    \newcommand{\todo}[1]{\textcolor{red}{\textbf{[TODO: #1]}}}
\else
    \newcommand{\karim}[1]{}
    \newcommand{\kavish}[1]{}
    \newcommand{\andrew}[1]{}
    \newcommand{\jy}[1]{}
  \newcommand{\todo}[1]{}
\fi

\title{WARP-RM: A Warp-Augmented Relative Progress Reward Model for Data Curation}

\author{
\bfseries
Justin Yu\textsuperscript{1 3 * \dag} \quad
Andrew Goldberg\textsuperscript{1 *} \quad
Kavish Kondap\textsuperscript{1 *} \quad
Karim El-Refai\textsuperscript{1 *} \\
\bfseries
Ethan Ransing\textsuperscript{1} \quad
Qianzhong Chen\textsuperscript{2} \quad
Mac Schwager\textsuperscript{2} \\
\bfseries
Fred Shentu\textsuperscript{3} \quad
Philipp Wu\textsuperscript{3} \quad
Ken Goldberg\textsuperscript{1} \\[4pt]
\textsuperscript{1}University of California, Berkeley \quad
\textsuperscript{2}Stanford University\quad
\textsuperscript{3}XDOF \\[3pt]
\textsuperscript{*}Equal contribution. \quad
\textsuperscript{\dag}Corresponding author: \texttt{yujustin@berkeley.edu}
}

\begin{document}
\maketitle
\hypersetup{pdftitle={WARP-RM: A Warp-Augmented Relative Progress Reward Model for Data Curation}, pdfauthor={Justin Yu, Andrew Goldberg, Kavish Kondap, Karim El-Refai, Ethan Ransing, Qianzhong Chen, Mac Schwager, Fred Shentu, Philipp Wu, Ken Goldberg}}

\begin{figure}[h]
\centering
\includegraphics[width=\textwidth]{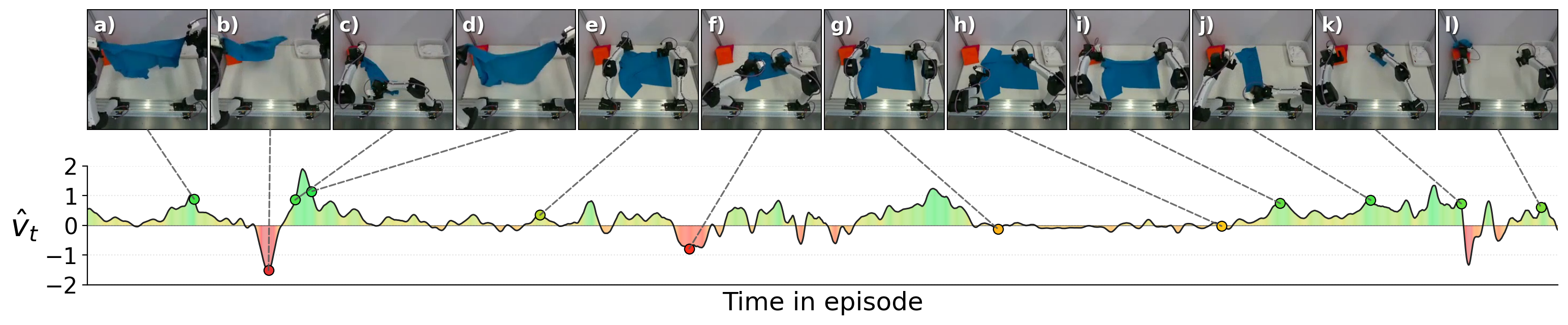}
\caption{\textbf{WARP-RM signed progress measure $\hat v_t$ on an unseen mixed-quality teleoperated T-shirt-folding demonstration.} Large negative magnitudes occur when the right gripper drops the shirt in \textbf{(b)}, and near-zero magnitude during stagnation between \textbf{(g)} and \textbf{(h)}. These values are used to filter and weight downstream policy training. Predictions on more examples in Appendix~\ref{app:qualitative}.}

\end{figure}

\begin{abstract}
Scaling imitation learning requires large datasets, yet human teleoperation inevitably produces mixed-quality demonstrations containing hesitations, retries, and pauses. Prior frame-level progress reward models supervise on absolute temporal progress proxies that suffer from label noise, or require costly human annotations to define subtask boundaries. We present \rewardmethod{} (Warp-Augmented Relative Progress), a novel fully self-supervised algorithm for learning dense, signed relative progress magnitudes directly from successful demonstrations. WARP generates per-frame progress targets via time-warp augmentations of demonstrations (variable playback speeds and reversals) and we train \progressmodel{} to predict normalized signed temporal displacement from the start of each sampled window. Aggregating these predictions across overlapping windows yields a dense frame-level progress signal. We then introduce \bcmethod{}, which uses these scalar reward estimates to filter and reweight action chunks during behavior cloning. We evaluate our approach on a physical bimanual robot system performing a long-horizon deformable object manipulation task: folding T-shirts from a random crumpled start. To evaluate policy robustness against suboptimal data, we construct training datasets of varying quality using episode length as a proxy for teleoperation sub-optimality. Across these training tiers, WARP-BC improves successful-folding throughput by up to $\sim\!\textbf{18}\times$ over vanilla BC. Furthermore, we evaluate bottle-in-bin placement in the real world and in simulation. Across 512 paired simulated scenes, WARP-BC achieves 290 bottles/hr versus 237 for vanilla BC and 271 for DemInf, with all curation methods retaining 31.5\% of the data. We release open simulation data, code, checkpoints, and evaluation artifacts for end-to-end reproduction of the WARP pipeline.

Project page: \url{https://uynitsuj.github.io/warp-rm/}. 
\end{abstract}


\section{Introduction}
Imitation learning has emerged as a powerful framework for long-horizon robot manipulation, enabling complex visuomotor behaviors from human teleoperated demonstrations. Recent advances in policy modeling and large-scale pretraining~\cite{zitkovich2023rt,black2026pi0visionlanguageactionflowmodel, chi2025diffusion, o2024open, team2024octo, huang2025otter,zhao2023learning, walke2023bridgedata,fu2024icrt} have significantly improved expressivity and generalization. However, these methods remain highly sensitive to demonstration quality, particularly in long-horizon manipulation settings, as policies learn to mimic suboptimal pauses and fumbles present in human teleoperation~\cite{mandlekar2021matters, beliaev2022imitation, brown2020better}. If trained on, these behaviors can derail policy performance. However, suboptimal sequences often contain valuable recovery behaviors, akin to DAgger data~\cite{ross2011reduction, liu2022robot, wu2025robocopilothumanintheloopinteractiveimitation, li2022efficient, kelly2019hg}. Some approaches to data curation operate at the trajectory level, discarding entire episodes that fall below a quality threshold~\cite{agia2025cupidcuratingdatarobot, hejna2025robot, lee2026quality}. However, this coarse filtering may suffer from two limitations: it discards valuable, high-advantage segments embedded within otherwise suboptimal executions, while simultaneously failing to prune localized hesitations or fumbles within retained demonstrations.

To address the limitations of episode-level filtering, recent methods instead learn frame-level progress signals for localized data curation. Most existing progress reward models operate in an \emph{absolute} progress regime. Methods like ReWiND~\cite{zhang2025rewind} use normalized episode duration as a supervision target, while representation learning techniques like VIP~\cite{ma2022vip} and LIV~\cite{ma2023liv} rely on temporal contrastive alignment to globally map visual observations. However, equating elapsed timesteps or global temporal alignment with task progress introduces considerable label noise. Because temporal progression does not guarantee task progression, two demonstrations at the same normalized frame index may reflect entirely different stages of the task due to pauses, failed grasps, or varying operator strategies~\cite{dwibedi2019temporal}.
More recent frame-level dense reward models address this noise with human annotations ~\cite{chen2026sarmstageawarerewardmodeling, mao2026arm}, but this is costly to scale and often inconsistent.


We propose \rewardmethod{}, a fully self-supervised method that learns a local \emph{relative} progress signal. We construct per-frame cumulative progress labels by replaying successful demonstrations at non-uniform velocities including in reverse. A model trained to predict these velocities from a sequence of images yields a frame-level progress-velocity signal of how fast and in which direction the task is advancing: large positive values during decisive forward progress, near zero during pauses or fumbles, and negative during state regression. Aggregating these predictions yields a dense progress signal that identifies decisive progress, stagnation, and regression. We then introduce \bcmethod{}, which uses these scores to filter and reweight action chunks during behavior cloning.

This paper makes four contributions:
\begin{enumerate}[topsep=2pt, itemsep=4pt, parsep=0pt, partopsep=0pt]
    \item WARP, a fully self-supervised algorithm that learns a dense relative task progress signal via a novel time-warping augmentation. We apply this method to train \progressmodel{}, a relative progress model that generates a signal indicating local task progress, given a demonstration.
    \item \bcmethod{}, which uses \progressmodel{}'s frame-level estimates
  to gate and reweight action chunks in the behavior cloning objective. We show (Sec.~\ref{sec:result}) that gating and reweighting by the terminal-frame progress velocity outperforms chunk-mean aggregation.
    \item We evaluate \bcmethod{} across \emph{420} real-world trials on a physical bimanual robot spanning two manipulation tasks. On T-shirt folding from a crumpled start, as the policy training dataset admits more inefficient demonstrations, vanilla BC collapses to a 2/20 success rate, while \bcmethod{} remains robust at 19/20 and yields up to an $\sim\!18\times$ improvement in successful folding throughput. Additionally, we validate our approach on a real-world bottle-in-bin placement task, where WARP-BC similarly improves execution speed and task throughput.
    \item Finally, we evaluate in a reproducible MuJoCo simulation of the bottle task built on the open-source ABC stack and dataset~\cite{abc2026}: across 512 paired scenes, \bcmethod{} achieves 290 bottles/hr versus 237 for vanilla BC and 271 for DemInf at matched retention across curation methods. We release open data, code, and evaluation artifacts for end-to-end reproduction of the WARP pipeline.

\end{enumerate}


\section{Related Work}
\label{sec:related_work}

\textbf{Time Warping for Video Representation Learning}. Predicting temporal transformations has been explored as a self-supervised objective for video representation learning. Many methods apply a uniform speed transformation and use speed classification as a self-supervised objective \cite{Yao_2020_CVPR, Wang20, chen2020RSPNet, huang2021ascnet}. A smaller body of work employs discrete, per-frame-independent warping, sampling the number of frames to skip at each timestep from a discrete set and training a classifier to predict the skip count per frame \cite{Dave_Jenni_Shah_2024, jenni2022videoretimelearningtemporallyvarying}. In contrast, \rewardmethod{} models time warping as a structured stochastic process that produces smoothly correlated, non-uniform speed variations spanning slow-motion to fast-forward playback. Rather than serving as a pretext task for representation learning, \rewardmethod{} uses time warping as an augmentation directly for its end task: progress estimation.

\textbf{Data Curation for Robot Imitation Learning}. 
Offline policy learning performance degrades substantially when training on mixed-quality human data containing variations in operator proficiency, trajectory length, and solution strategy \cite{mandlekar2021matters}. Prior work has explored demonstration data curation at varying levels of granularity to address sensitivity to suboptimal demonstrations. At the dataset level, Re-Mix~\cite{hejna2024remixoptimizingdatamixtures} optimizes over data sources to create large multi-dataset mixtures.
At the episode level, DemInf~\cite{hejna2025robot} uses a mutual information approach to curate episodes, while other works use influence function based approaches \cite{agia2025cupidcuratingdatarobot, lee2026quality}.


\textbf{Progress Reward Models}. Rather than operating at the episode level, reward and progress models learn a dense frame-level quality signal within a demonstration, enabling finer-grained data curation. Most prior progress models operate in an \emph{absolute} progress regime: each
frame is assigned a target value on a globally normalized $[0,1]$ completion axis, either from its position within the demonstration or from human-annotated subtask boundaries. ReWiND~\cite{zhang2025rewind} supervises on normalized frame indices, while VIP~\cite{ma2022vip} and LIV~\cite{ma2023liv} use temporal contrastive learning to align representations with absolute task progression. We argue that this becomes a limitation for long-horizon teleoperation. Operators may pause, retry, or recover at different points, thus the same normalized timestamp can correspond to substantially different task states across demonstrations, injecting noise into any model that supervises against the absolute axis. In contrast, \rewardmethod{} models \emph{relative progress velocity}: whether a short temporal segment moves the task forward or backward, and by how much. This formulation avoids requiring cross-demonstration temporal alignment while still producing a dense scalar signal suitable for action-chunk reweighting.

Closest to our work, SARM~\cite{chen2026sarmstageawarerewardmodeling} and ARM~\cite{mao2026arm} leverage human annotated data to train progress reward models and use the learned models to reweigh action chunks, performing Reward-Aligned Behavior Cloning (RA-BC). SCIZOR~\cite{zhang2025scizor} trains a self-supervised progress estimator which predicts the time between two frames to curate large multi-task datasets for behavior cloning. \bcmethod{} additionally weights retained chunks in proportion to their magnitude. 
Other works~\cite{liang2026robometer, chen2026topreward, tan2025robo, liang2023holistic} explore multi-task reward modeling using large vision-language models (VLMs). While these methods demonstrate impressive semantic generalization, they often trade high-frequency temporal resolution for this breadth. For instance, methods relying on token-probability extraction~\cite{chen2026topreward} or trajectory-level preference comparisons~\cite{liang2026robometer} require VLM queries that necessitate coarser frame subsampling, limiting sensitivity to higher frequency temporal variations critical for dynamic control. Furthermore, because these models frequently anchor on absolute progress~\cite{liang2026robometer} or discretized semantic milestones~\cite{tan2025robo}, they remain susceptible to cross-demonstration temporal alignment noise.

\section{Method}
  \label{sec:method}

\subsection{Overview and Notation}
We introduce the Warp-Augmented Relative Progress Reward Model (WARP-RM), a vision-based model that estimates the dense, per-frame progress velocity of a task directly from visual observations. To train WARP-RM without human annotations, we propose the WARP algorithm, a fully self-supervised method that trains the model to predict the normalized time delta for each frame relative to a starting frame, using a dataset of successful human teleoperated demonstrations. We hypothesize that time-warping video playback speed—and its corresponding progress label—provides a strong self-supervised signal: subsampling frames at wider intervals simulates faster execution, while sampling at narrower intervals simulates slower execution. By varying this time-warping augmentation non-uniformly, a single demonstration provides dense supervision over a continuous range of progress velocities.


\label{sec:method_overview}

A demonstration consists of a sequence of $T$ RGB frames $o_0, \dots, o_{T-1}$ recorded at a fixed frequency $f$, from which a frozen visual encoder $\phi$ produces a per-frame feature vector. From each demonstration we sample a window of $N$ frame indices $i_0, \dots, i_{N-1}$ via a time-warping procedure that replays the demonstration at varying speeds and directions, so the indices may be non-monotonic and non-linear. Each index has a corresponding pseudo-label $y_k$: the normalized time delta from $o_{i_0}$ to $o_{i_k}$, representing the cumulative temporal displacement from the start of the window.

\progressmodel{} is trained to predict $y_1,...y_{N-1}$ given the set of visual features $\phi(o_{i_0}), ..., \phi(o_{i_{N-1}})$. At inference the model is given a sequence of frames spaced linearly with a canonical inter-frame stride of $S$ seconds covering a window of $L = (N-1)S$ seconds, and predictions are aggregated following the procedure in Sec.~\ref{sec:method_warp_bc}. This assigns each frame a scalar \textit{signed progress velocity} $v_t$ measuring how fast and in which direction the task is progressing, calibrated so that: $\hat{v}_t \approx 1$ matches the pace of the average reference demonstrations, $\hat{v}_t \approx 0$ is stalling progress, and $\hat{v}_t < 0$ is regressing progress.

\subsection{Time-Warp Sampler}
  \label{sec:method_warp}

  \begin{figure}[h]
\centering
\includegraphics[width=0.90\textwidth]{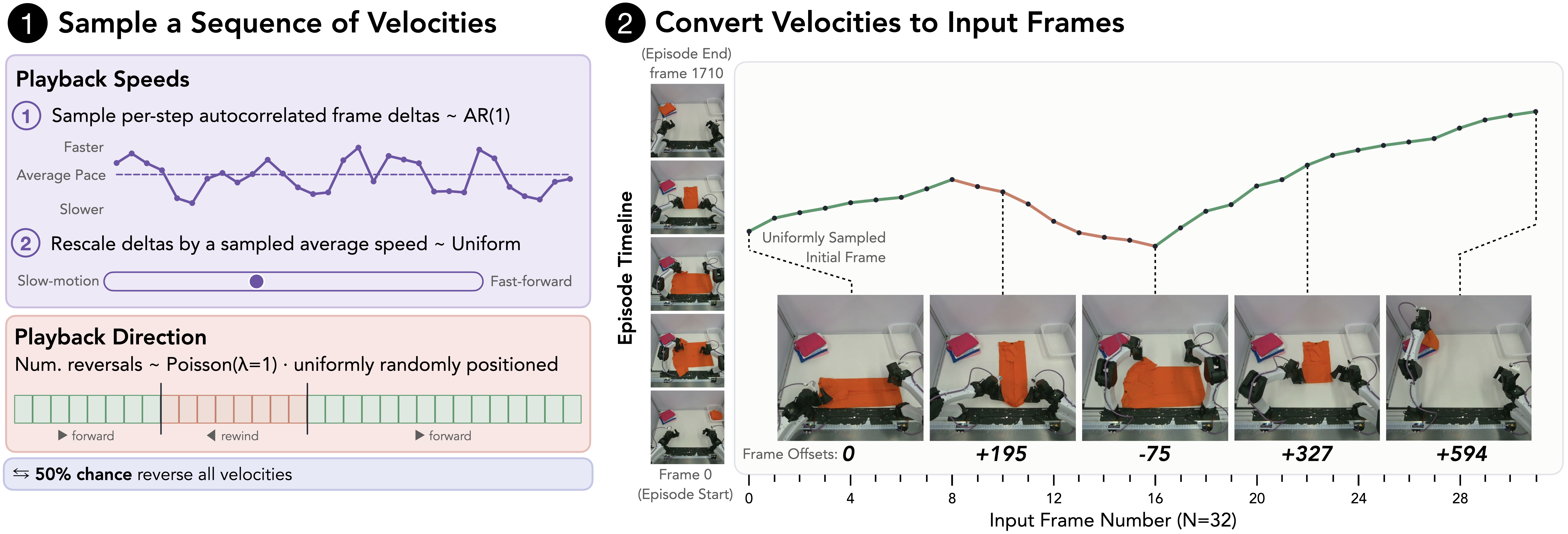}
\caption{\textbf{Time-Warp Sampler.}
WARP resamples trajectories using a warped playback schedule. (\textbf{1}): Playback speed varies to span slow-motion to fast-forward. Playback direction is randomly inverted to expose the model to negative progress (regression). (\textbf{2}): Accumulating these playback speeds yields a window of $32$ source frames. The relative offset of each frame from the starting frame serves as the self-supervised progress label for the WARP-RM model to predict.}
\end{figure}

The sampler produces a set of indices $i_0,...i_{N-1}$ through the following procedure. First, $N-1$ relative log-velocities are drawn from a stationary AR(1) process with autocorrelation $\alpha$ and marginal standard deviation $\sigma_\infty$:
\begin{equation}
    z_0 \sim \mathcal{N}(0,\,\sigma_\infty^2), \qquad
    z_k = \alpha\, z_{k-1} + \sqrt{1-\alpha^2}\,\sigma_\infty\,\epsilon_k,\quad
    \epsilon_k \sim \mathcal{N}(0,\,1),\quad k = 1, \dots, N-2
\label{eq:ar1}
\end{equation}

Exponentiating the log-velocities gives positive playback speeds $\tilde v_k=e^{z_k}$. Log-space sampling treats a $2\times$ speedup and a $0.5\times$ slowdown symmetrically, while $\alpha$ controls the smoothness of successive speed changes. We sample a path-length budget $\ell$ uniformly from $[fL/3,5fL/3]$ source frames, cap it at $T-1$, and rescale the speeds:
\begin{equation}
\tilde u_k=\ell\,\tilde v_k\big/{\textstyle\sum_{j=0}^{N-2}\tilde v_j},\qquad k=0,\ldots,N-2.
\label{eq:rescale}
\end{equation}
Following ReWiND~\cite{zhang2025rewind}, reversed playback supplies negative-displacement examples. We draw $R=\min(R_0,N-2)$ reversal points, with $R_0\sim\mathrm{Poisson}(\lambda_{\mathrm{rev}})$, uniformly without replacement from $\{1,\ldots,N-2\}$. Each reversal flips the sign of subsequent increments; all signs are additionally flipped with probability $0.5$. The resulting signed increments $u_k$ accumulate to $c_j=\sum_{k=0}^{j-1}u_k$, with $c_0=0$. Sampling a valid starting frame $i_0$ then gives $i_j=\mathrm{round}(i_0+c_j)$. Reversals can make the visited span smaller than $\ell$. Simulation retains this absolute path budget with a shorter label-normalization stride (App.~\ref{app:sim_protocol}).

\ifcorlcameraready
\else
Given the cumulative signed displacements $c_j$, draw $i_0$ uniformly from the integer range $[\lceil-\min_j c_j\rceil,\lfloor T-1-\max_j c_j\rfloor]$. These bounds ensure that every sampled index lies within $[0,T-1]$.

\fi

  \subsection{Progress Model Training}
  \label{sec:method_progress}

  \textbf{Relative Cumulative Progress Targets.} From the sampled indices, the per-frame cumulative progress pseudo-labels are computed as the signed displacement from the window start normalized by $C_{\text{norm}}=f(N-1)S$:
  \begin{equation}
    y_j \;=\; (i_j - i_0) / C_{\text{norm}}, \qquad j = 0, \dots, N{-}1
    \label{eq:rel_label}
\end{equation}

\textbf{Categorical Objective.} Given the visual features $\phi(o_{i_0}), ..., \phi(o_{i_{N-1}})$ the progress model is trained to predict $y_0, ..., y_{N-1}$. Rather than regressing directly on $y_j$, \progressmodel{} predicts $N$ separate probability distributions over evenly spaced categorical bins. The model is trained with a cross-entropy loss against a two-hot target encoding that encodes a continuous value by finding the coefficients for linear interpolation from the two nearest bin centers. This categorical classification approach mitigates optimization instabilities and capacity underutilization typically found in direct regression \cite{Farebrotheretal2024}.

  \subsection{Model Architecture}
  \label{sec:method_arch}

\begin{figure}[h]
\centering
\includegraphics[width=0.65\textwidth]{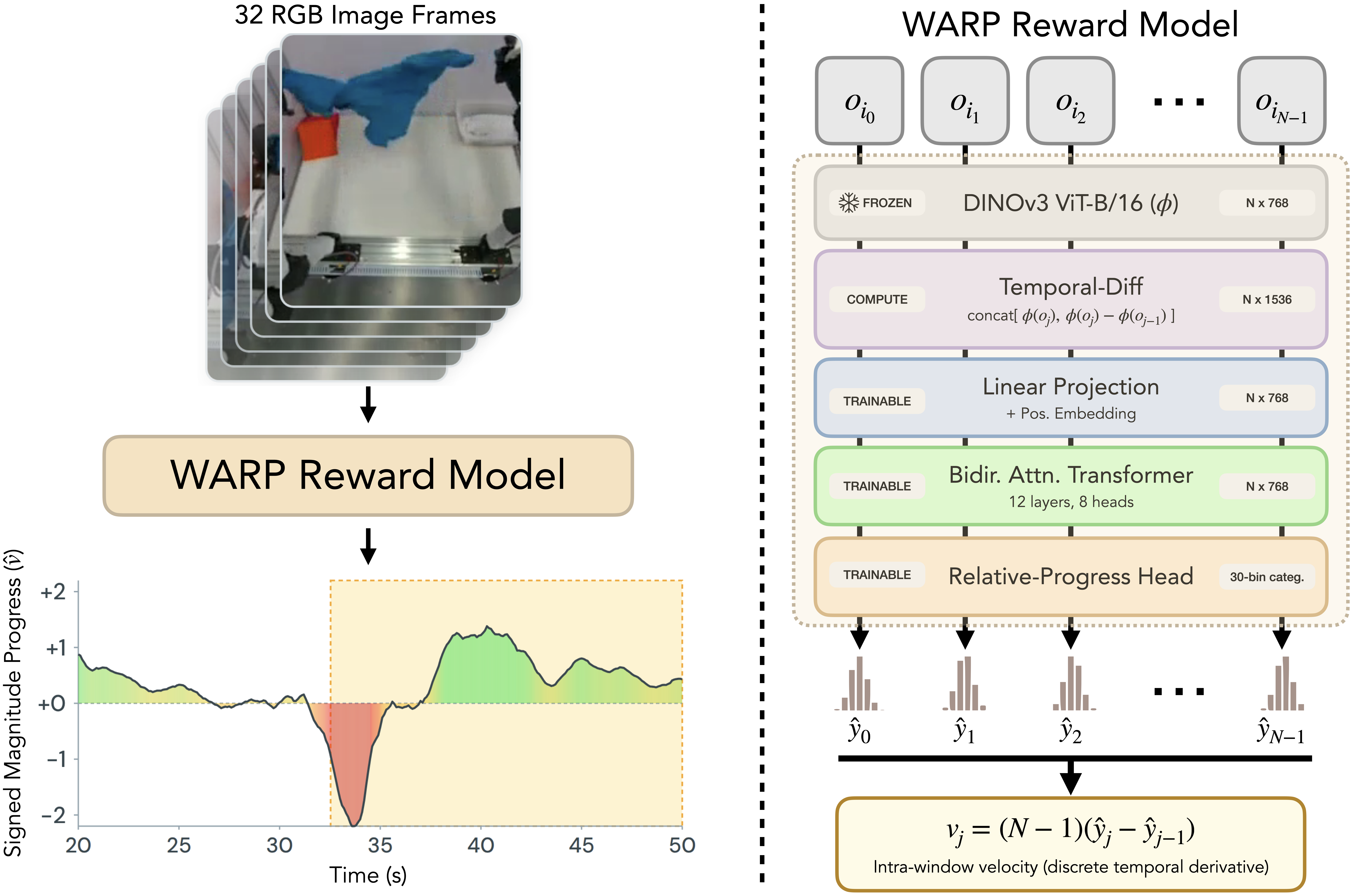}
\caption{\textbf{\rewardmethod{}-RM Architecture.}
    A 32-frame demonstration window (left) is encoded by a frozen DINOv3 backbone $\phi$ and aggregated by a bidirectional-attention transformer that emits a distribution over 30 cumulative-progress bins at each input frame. The yellow shaded region (bottom-left) illustrates one such sliding prediction window applied to the continuous episode. Their per-frame expectations form the window's predicted cumulative progress vector $\hat{y} \in \mathbb{R}^N$. A discrete temporal derivative yields intra-window velocities $v_j$ (bottom right), which are then averaged across overlapping windows to produce $\hat{v}_t$. See Sections ~\ref{sec:method_arch},~\ref{sec:method_warp_bc} for details.
  \label{fig:warp_architecture}}
\end{figure}

  Similar to SARM \cite{chen2026sarmstageawarerewardmodeling}, the model architecture uses a frozen visual backbone and transformer temporal aggregator, but it replaces the stage classifier and subtask-progress regressor with a single progress velocity head (Figure~\ref{fig:warp_architecture}). \ifcorlcameraready
\else
WARP-RM uses a frozen DINOv3 ViT-B/16~\cite{simeoni2025dinov3} as $\phi$, yielding a $768$-dimensional feature vector per frame. 
  Each input token is formed by concatenating the frame embedding with its temporal difference: $[\phi(o_{i_j}),\, \phi(o_{i_j}) - \phi(o_{i_{j-1}})] \in \mathbb{R}^{1536}$, with the difference set to zero for $j=0$. Tokens are projected to the transformer dimension, augmented with fixed sinusoidal positional embeddings, and processed by a bidirectional Transformer encoder. Finally, a linear layer maps each output token to separate probability distributions. At inference, the predicted $\hat y_j$ is computed as the expectation of the predicted distribution.

\fi

\subsection{Warp-Augmented Relative Progress Behavior Cloning}
  \label{sec:method_warp_bc}

  \bcmethod{} aggregates predictions from WARP-RM into dense, per-frame progress velocities and uses them for binary filtering or continuous reweighting of action chunks.

  \textbf{Velocity Aggregation.} 
  We apply \progressmodel{} in overlapping windows, each containing $N$ frames separated by the canonical stride of $S$ seconds, with consecutive windows shifted by a single source-frame. Within each window, consecutive expected cumulative progress predictions are differenced and scaled to produce a sequence of intra-window velocities: $v_j = (N-1)(\hat{y}_j - \hat{y}_{j-1})$. Each interior source-frame $t$ in the episode is covered by multiple overlapping windows. The final per-frame progress velocity, $\hat{v}_t$, is computed as the mean of all intra-window velocity predictions $v_j$.
  

  \textbf{Action Chunk Weighting.} The behavior cloning policy is trained to predict one second action chunks. The velocity at the final frame of the action chunk, $\hat v_{\text{end}}$, is used to compute the continuous reward-aligned behavior cloning (RA-BC) weight $w(s, a)$ for each state-action pair:
  \begin{equation}
  w(s, a) \;=\; \hat{v}_{\text{end}} \, \cdot \, \mathbf{1}_{\, \hat{v}_{\text{end}} \,>\, \tau \,}
  \label{eq:warpbc_weight}
  \end{equation}
The binary variant uses the same threshold but assigns unit weight to retained chunks: $w(s,a)=\mathbf{1}_{\hat v_{\text{end}}>\tau}$. Chunks with $w(s, a) = 0$ are filtered out \emph{prior} to training to avoid reducing effective batch size, following advantage-filtered behavior cloning~\cite{grigsby2023closerlookadvantagefilteredbehavioral}. Here $\hat v_{\text{end}}$ acts as an empirical progress velocity-based proxy for advantage rather than an explicitly estimated RL advantage with a value baseline.

  \textbf{Behavior Cloning Objective.} The final policy loss is the standard flow-matching loss weighted per-sample by $w(s, a)$:
  \begin{equation}
  \mathcal{L}_\text{BC} \;=\; \mathbb{E}_{(s, a) \sim \mathcal{D}}\, [w{(s,a)} \cdot \mathcal{L}_\text{flow}(\pi_\theta; s, a)].
  \label{eq:bc_loss}
  \end{equation}

\newsavebox{\lefttablebox}
\newsavebox{\righttablebox}

\savebox{\lefttablebox}{%
  \small\setlength{\tabcolsep}{3pt}\renewcommand{\arraystretch}{1.10}%
  \begin{tabular}[t]{ll rrr}
    \toprule
    \multicolumn{2}{l}{} & \multicolumn{3}{c}{\textbf{Dataset / Tier}} \\
    \cmidrule(lr){3-5}
    \textbf{Method} & \textbf{Metric} & $\mathcal{D}_1$ & $\mathcal{D}_2$ & $\mathcal{D}_3$ \\
    \midrule
    \multirow{3}{*}{Vanilla BC}
      & Success $\uparrow$            & \textbf{20/20} & 2/20  & 0/20 \\
      & Mean TTC (s) $\downarrow$     & 113.8 & 199.0 & N/A  \\
      & Thrput (/hr) $\uparrow$       & 31.6  & 1.5  & 0.0  \\
      & Act. Chunks Kept & 100\% & 100\% & 100\% \\
    \midrule
    \multirow{3}{*}{\textbf{\bcmethod{}}}
      & Success $\uparrow$            & \textbf{20/20} & \textbf{19/20} & \textbf{14/20} \\
      & Mean TTC (s) $\downarrow$     & \textbf{63.9} & \textbf{118.8} & \textbf{117.4} \\
      & Thrput (/hr) $\uparrow$       & \textbf{56.3} & \textbf{27.4}  & \textbf{16.3} \\
      & Act. Chunks Kept & 35.7\% & 34.4\% & 22.5\% \\
    \bottomrule
  \end{tabular}%
}

\savebox{\righttablebox}{%
    \small\setlength{\tabcolsep}{3pt}\renewcommand{\arraystretch}{1.10}%
    \begin{tabular}[t]{ll rr}
      \toprule
      \multicolumn{2}{l}{} & \multicolumn{2}{c}{\textbf{Dataset / Tier}} \\
      \cmidrule(lr){3-4}
      \textbf{Method} & \textbf{Metric} & $\mathcal{D}_4$ & $\mathcal{D}_5$ \\
      \midrule
      \multirow{4}{*}{SARM~\cite{chen2026sarmstageawarerewardmodeling}}
        & Success $\uparrow$            & 19/20 & 2/20 \\
        & Mean TTC (s) $\downarrow$     & 90.5  & 156.0 \\
        & Thrput (/hr) $\uparrow$       & 34.9  & 1.55 \\
        & Act. Chunks Kept                     & 78.5\% & 66.6\% \\
      \midrule
      \multirow{4}{*}{DemInf~\cite{hejna2025robot}}
        & Success $\uparrow$            & 19/20 & 18/20 \\ 
        & Mean TTC (s) $\downarrow$     & 89.6 & 115.8 \\ 
        & Thrput (/hr) $\uparrow$       & 35.2 & 25.3  \\
        & Act. Chunks Kept                    & 45.6\% & 33.7\% \\
      \midrule
      \multirow{4}{*}{SCIZOR~\cite{zhang2025scizor}}
        & Success $\uparrow$            & 19/20 & 2/20 \\ 
        & Mean TTC (s) $\downarrow$     & 98.4 & 206.2 \\ 
        & Thrput (/hr) $\uparrow$       & 32.4 & 1.5 \\
        & Act. Chunks Kept                     & 77.9\% & 66.7\% \\
      \midrule
      \multirow{4}{*}{\textbf{\bcmethod{}}}
        & Success $\uparrow$            & \textbf{20/20} & \textbf{20/20} \\
        & Mean TTC (s) $\downarrow$     & \textbf{71.2} & \textbf{80.7} \\
        & Thrput (/hr) $\uparrow$       & \textbf{50.6} & \textbf{44.6} \\
        & Act. Chunks Kept                     & 45.6\% & 33.7\% \\
      \bottomrule
    \end{tabular}%
}

\begin{table}[t!]
\begin{minipage}[t]{\wd\lefttablebox}
  \usebox{\lefttablebox}
  \vspace{3pt}
  \captionof{table}{\textbf{Cross-tier evaluation} on bimanual \\ T-shirt folding from a crumpled start compared across three training datasets that admit progressively more suboptimal demonstrations. We report success rates, mean time-to-completion (TTC), throughput, and \% of action chunks kept for training. WARP-BC sustains higher throughput across tiers, whereas vanilla BC degrades as broader demonstrations are admitted into training.}
  \label{tab:main_results}
\end{minipage}%
\hspace{5pt}%
\begin{minipage}[t]{\wd\righttablebox}
  \usebox{\righttablebox}
  \vspace{3pt}
  \captionof{table}{\textbf{Matched baseline comparisons} on $\mathcal{D}_4 = \mathcal{D}_1 \cup \mathcal{D}_A$, $\mathcal{D}_5 = \mathcal{D}_2 \cup \mathcal{D}_A$.}
  \label{tab:sarm_compare}
\end{minipage}%
\end{table}

\section{Experiments} \label{sec:result}

We evaluate in three settings: real-world T-shirt folding, real-world bottle-in-bin placement, and a reproducible simulated bottle-in-bin benchmark. In the T-shirt setting, each policy is evaluated on a bimanual I2RT YAM robot arm setup over $20$ trials\footnote{At $n=20$, single-trial success differences (e.g., 19/20 vs.\ 20/20) fall within binomial sampling noise; only larger gaps should be read as meaningful. Where success rates are comparable, throughput and time-to-completion are more informative metrics.} of T-shirt folding using a variety of differently colored medium-sized T-shirts not seen during training. In each trial, the robot must retrieve a crumpled shirt from a bin containing a variable number of shirts, flatten it on the workspace, fold both sleeves inward, fold the shirt in half twice, and move it towards the top-left corner of the workspace to complete the task. If the policy does not complete the task within
240 seconds, the trial is marked as a failure. This time budget is chosen as 2$\times$ the longest training demonstration, providing a generous margin for retry behaviors while bounding evaluation time. Time-to-completion (TTC) is reported as the average completion time for successful trials. Throughput is the number of successful folds per hour, with each failed trial contributing its four-minute timeout to the denominator; human reset time between trials is excluded from this calculation.
\subsection{Implementation Details}

We use $N=32$ frames, stride $S=1.5s$, AR(1) parameters $\alpha=0.5, \sigma_{\infty}=\ln 2$, reversal rate $\lambda_{rev}=1$, and filtering threshold $\tau=1.0$ (reweighting only chunks above the reference expert pace, $\hat v_{\text{end}} > 1$). See Appendix~\ref{app:impl} for additional details.

\subsection{Policy Training Datasets and Tiers}
We form three length-filtered tiers from one pool of successful teleoperated T-shirt demonstrations: $\mathcal{D}_1$ ($\le60$\,s), $\mathcal{D}_2$ ($\le90$\,s), and $\mathcal{D}_3$ ($\le120$\,s). Longer episodes tend to contain more hesitations and retries, providing a coarse test of robustness to inefficient behavior. Vanilla BC and WARP-BC train on the same demonstrations within each tier. One WARP-RM, trained on the fixed set of 1,950 shortest demonstrations ($\mathcal{D}_{RM}$, $\le59.8$\,s), scores all tiers. Ablations use $\mathcal{D}_2$; counts and distributions appear in App.~\ref{app:data}.
\ifcorlcameraready\else
The tiers contain 2,427 episodes (36.1 hours), 4,124 (71.3 hours), and 6,473 (139.7 hours), respectively. Episode lengths have a dominant mode near 50--60 seconds and a broader tail beyond 85 seconds.
\fi

\subsection{Cross-tier Results}

\ifcorlcameraready\else
\begin{figure}[t!]
    
  \centering
  \includegraphics[width=0.75\linewidth]{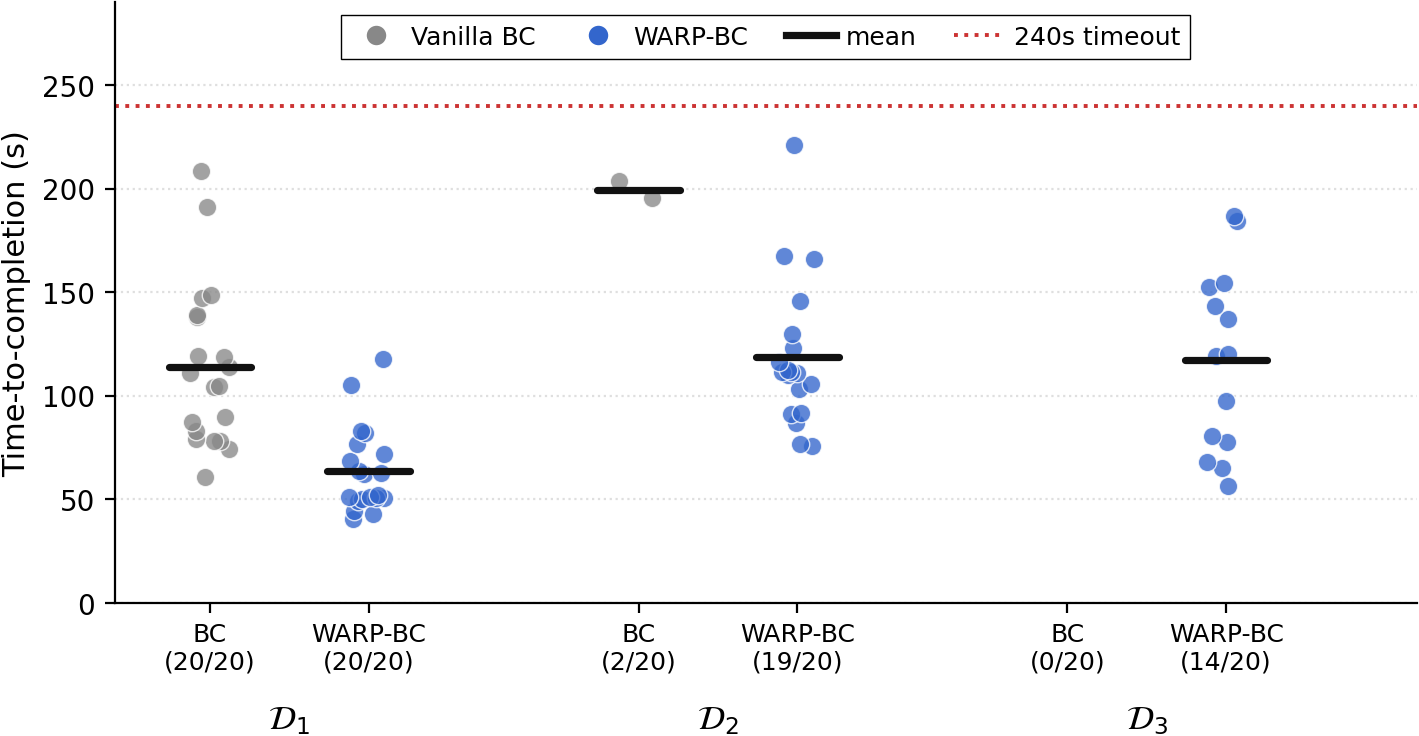}
  
  \caption{\textbf{Time-to-completion distribution for successes.} Performance is evaluated across three datasets tiered by increasing demonstration sub-optimality: $\mathcal{D}_1$ ($\le 60$s, efficient demonstrations), $\mathcal{D}_2$ ($\le 90$s, moderate inefficiencies), and $\mathcal{D}_3$ ($\le 120$s, demonstrations with more operator hesitations and recoveries). Policy rollouts which exceed 240 seconds are considered failures and are not shown.}
  \label{fig:ttc_distribution}
\end{figure}

\fi

As shown in Table~\ref{tab:main_results} and Figure~\ref{fig:ttc_distribution}, while both methods succeed on $\mathcal{D}_{1}$, WARP-BC yields a 1.78$\times$ throughput improvement, with performance divergence increasing on $\mathcal{D}_{2}$ ($\sim18\times$ throughput improvement  compared to vanilla BC) and $\mathcal{D}_{3}$. From qualitative observation, vanilla BC on the later tiers often gets caught in repetitive, localized adjustments that fail to progress the task, often cycling through micro-adjustment behavior until the 240-second timeout is reached.

\subsection{Baseline Comparisons} 
Because SARM requires stage annotations, all methods use augmented datasets $\mathcal{D}_4=\mathcal{D}_1\cup\mathcal{D}_A$ and $\mathcal{D}_5=\mathcal{D}_2\cup\mathcal{D}_A$; other methods treat $\mathcal{D}_A$ as unlabeled. On the broader tier, SARM and SCIZOR both fall from 19/20 to 2/20 successes, while DemInf retains 18/20 and WARP-BC achieves 20/20 (Table~\ref{tab:sarm_compare}). WARP-BC also yields the highest throughput on both tiers. WARP-BC and DemInf have matched retention budgets, whereas SARM and SCIZOR retain more data. The simulation comparison below controls this difference across all curation methods. Completion-time distributions appear in App.~\ref{app:exp_statistics}.

\subsection{Ablations}

\begin{table}[t]
    \centering
    \small
    \setlength{\tabcolsep}{2pt}
    \renewcommand{\arraystretch}{1.1}
    \begin{tabular}{ll rrrr}
    \toprule
    \textbf{Ablations} & \textbf{Variant} & \textbf{Success} & \textbf{Mean TTC (s)} & \textbf{Thrput (/hr)} & \textbf{Act. Chunks Kept} \\
    \midrule
    \multirow{3}{*}{Weighting}
      & $\tau = 0$                              & 3/20           & 201.4          & 2.3           & 97.0\% \\
      & $\tau = 1$, binary                      & 16/20          & 139.6          & 18.0          & 34.4\% \\
      & $\tau = 1$, continuous \textbf{[\bcmethod{}]}  & \textbf{19/20} & \textbf{118.8} & \textbf{27.4} & 34.4\% \\
    \midrule
    \multirow{3}{*}{Aggregation}
      & Mean $\hat v$ over chunk                & 15/20          & 127.0          & 17.4          & 34.0\% \\
      & Mean $\hat v$ over chunk, future offset & 14/20          & 124.2          & 15.9          & 34.3\% \\
      & Terminal $\hat v_\text{end}$ \textbf{[\bcmethod{}]} & \textbf{19/20} & \textbf{118.8} & \textbf{27.4} & 34.4\% \\
    \midrule
    \multirow{2}{*}{Sampler}
      & IID log-normal                          & 18/20          & 131.0          & 22.8          & 28.7\% \\
      & AR(1) process \textbf{[\bcmethod{}]}     & \textbf{19/20} & \textbf{118.8} & \textbf{27.4} & 34.4\% \\
    \bottomrule
    \end{tabular}
    \vspace{4pt}
    \caption{\textbf{Ablations} on $\mathcal{D}_2$. Each row varies an algorithmic component
    while holding others fixed. We investigate: 
(i) \emph{Weighting Function} (Eq.~\ref{eq:warpbc_weight}), how scores are filtered or reweighted in policy training loss,
(ii) \emph{Aggregation Strategy}, how frame-level progress is summarized over an action chunk, and 
(iii) \emph{Time-Warp Sampler}, comparing temporally independent vs. smoothly correlated AR(1) sampling.
We report success rate, mean time-to-completion (TTC) for successful rollouts, throughput, and the percent of action chunks kept for policy training.}
    \label{tab:ablations}
\end{table}

\textbf{Weighting and aggregation.} Binary and continuous weighting retain the same 34.4\% of chunks, with 16/20 and 19/20 successes, respectively. Terminal-velocity aggregation yields 19/20 successes, compared with 15/20 for the aligned chunk mean and 14/20 for the future-offset mean (Table~\ref{tab:ablations}). These small-sample differences are descriptive.

\textbf{Time-warp sampler.} AR(1) and IID sampling yield similar success counts (19/20 vs.\ 18/20), while throughput is 27.4 vs.\ 22.8 folds/hr. Retention also differs (34.4\% vs.\ 28.7\%), so this comparison does not isolate temporal correlation from data selection. The simulation ablation below compares AR(1) and IID at matched retention.

\subsection{Bottle-in-Bin Placement Task in the Real World}
\label{sec:bottles}
\begin{table}[t]
  \centering
  \small
  \setlength{\tabcolsep}{6pt}
  \renewcommand{\arraystretch}{1.15}
  \begin{tabular}{l ccccc}
    \toprule
    \textbf{Method}
      & \makecell{\textbf{Bottles}\\\textbf{Placed} $\uparrow$}
      & \makecell{\textbf{Mean Time /}\\\textbf{Bottle (s)} $\downarrow$}
      & \makecell{\textbf{Thrput}\\\textbf{(/hr)} $\uparrow$}
      & \makecell{\textbf{Act. Chunks}\\\textbf{Kept}} \\
    \midrule
    Vanilla BC
      & 59/80 & 15.9 & 147.8 & 100\% \\
    \textbf{\bcmethod{}}
      & \textbf{74/80} & \textbf{11.3} & \textbf{237.8} & 30.6\% \\
    \bottomrule
  \end{tabular}
  \vspace{4pt}
  \caption{\textbf{Real-world bottle-in-bin placement.} Each policy runs 20 four-bottle trials with a 90\,s timeout (80 bottles total). Placement time averages intervals between successful drops; throughput counts placed bottles per hour and charges timed-out trials the full 90\,s.}
  \label{tab:bottle_results}
\end{table}

\ifcorlcameraready\else
\begin{figure}[t!]
  \centering
  \includegraphics[width=0.5\linewidth]{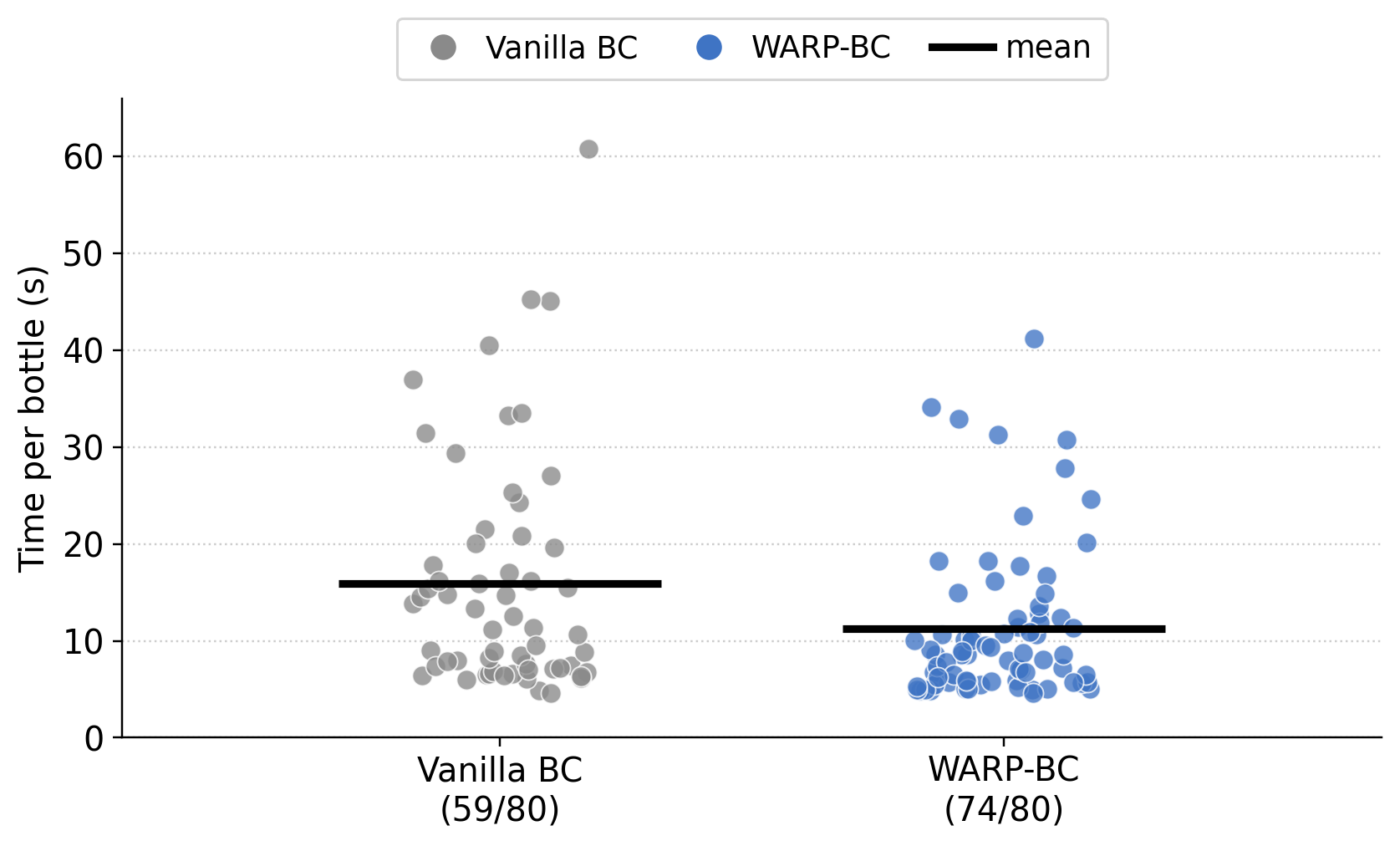}
  
  \caption{\textbf{Per-bottle placement-time distribution for the bottle-in-bin task.}
  Each point is the time to place a single bottle (interval between consecutive drops);
  gray points are vanilla BC ($59$ bottles placed) and blue are \bcmethod{}
  ($74$ bottles placed), with the total placed out of $80$ shown under each label. Black
  bars denote the mean ($15.9$\,s vs.\ $11.3$\,s). \bcmethod{} places bottles faster and
  with a tighter distribution, while vanilla BC exhibits a heavier tail of slow placements.}
  \label{fig:bottle_tpb_distribution}
\end{figure}

\fi
Beyond T-shirt folding, we evaluate \bcmethod{} on a bottle-in-bin placement task, where the
robot places four plastic bottles into a bin ($90$\,s timeout, $20$ trials per method).
Both policies train on the same dataset of bottle demonstrations, and \progressmodel{} uses
hyperparameters identical to the T-shirt setting (App.~\ref{app:bottle_data}).
Table~\ref{tab:bottle_results} and Figure~\ref{fig:bottle_tpb_distribution} show \bcmethod{}
places $74/80$ bottles versus $59/80$ for vanilla BC, reducing mean per-bottle placement time
from $15.9$ to $11.3$\,s and improving throughput by $1.6\times$ ($237.8$ vs.\ $147.8$
bottles/hr). Mean per-bottle placement time measures execution speed over successfully placed bottles only and is not penalized by missed bottles, whereas throughput charges each timed-out trial its full 90 s timeout and thus additionally reflects task completion. See App.~\ref{app:eval_protocol} for extended evaluation-protocol details.

\subsection{Bottle-in-Bin Placement Task in Simulation}
\label{sec:sim_bottles}
\begin{table}[t]
  \centering
  \small
  \begin{tabular}{lrrrr}
    \toprule
    \textbf{Method} & \textbf{Data kept} & \textbf{Bottles/scene} $\uparrow$ & \textbf{Thrpt (/hr)} $\uparrow$ & \textbf{All 6} $\uparrow$ \\
    \midrule
    Vanilla BC & 100\% & 3.885 & 237 & 9.4\% \\
    Random & 31.5\% & 3.770 & 230 & 10.9\% \\
    ReWiND & 31.5\% & 3.781 & 231 & 9.4\% \\
    SARM (oracle) & 31.5\% & 4.191 & 265 & 20.5\% \\
    WARP-BC (IID) & 31.5\% & 4.285 & 269 & 18.9\% \\
    SCIZOR & 31.5\% & 4.299 & 270 & 19.1\% \\
    DemInf & 31.5\% & 4.332 & 271 & 18.8\% \\
    \textbf{\bcmethod{}} & 31.5\% & \textbf{4.533} & \textbf{290} & \textbf{25.0\%} \\
    \bottomrule
  \end{tabular}
  \caption{\textbf{Simulated bottle-in-bin results.} All methods use 512 paired six-bottle scenes (nominal 60\,s horizon). Curation methods retain 31.5\% of the data. Both WARP-BC variants use binary filtering; SARM uses oracle stage boundaries. ``All 6'' is the fraction of scenes with every bottle placed.}
  \label{tab:sim_bottle_results}
\end{table}

\ifcorlcameraready\else
\begin{figure}[ht]
  \centering
  \includegraphics[width=\linewidth]{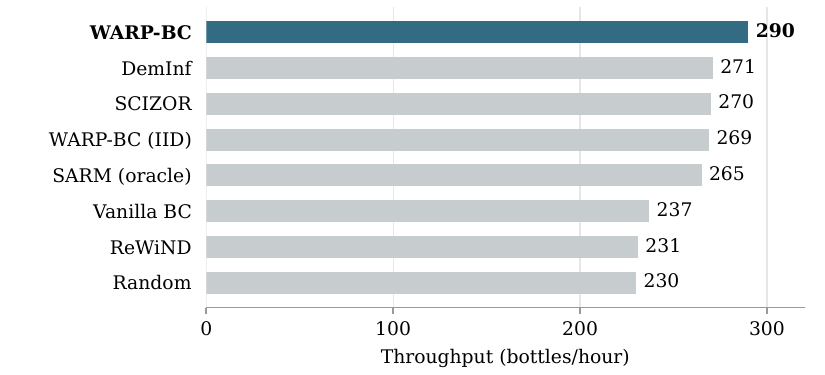}
  \caption{\textbf{Simulation throughput across 512 paired scenes.} WARP-BC achieves 290 bottles/hr, compared with 271 for DemInf and 237 for vanilla BC. All curation methods retain 31.5\% of the data; vanilla BC uses 100\%. Paired confidence intervals are reported in the text.}
  \label{fig:sim_bottle_comparison}
\end{figure}

\fi
To extend the hardware comparison to a larger evaluation, we use the public ABC MuJoCo-Warp environment~\cite{abc2026} and 2,438 demonstrations on the bimanual YAM platform. Each method is evaluated on the same 512 six-bottle scenes with a nominal 60\,s horizon. Policies share the demonstrations, $\pi_0$ architecture, and training-step budget. All curation methods retain 31.5\% of the data; vanilla BC uses the full dataset. WARP-BC uses binary filtering ($\tau=1$), giving retained chunks equal loss weight. SARM receives oracle stage boundaries from simulator state, since human stage annotations are unavailable. Further details are in App.~\ref{app:sim_protocol}.

WARP-BC achieves the highest throughput among the eight methods (Table~\ref{tab:sim_bottle_results}), placing 4.533 bottles/scene at 290 bottles/hr, compared with 3.885 and 237 for vanilla BC. The improvement also translates into more completed scenes: WARP-BC clears all six bottles in 25.0\% of scenes, versus 9.4\% for vanilla BC. Its paired throughput gain is +53 bottles/hr over vanilla (95\% CI [42, 64]), with gains of +19 over DemInf ([9, 29]) and +20 over SCIZOR ([8, 32]). It also places 0.342 more bottles/scene than oracle-SARM ([0.18, 0.51]). Figure~\ref{fig:sim_bottle_comparison} shows the throughput comparison.

\textbf{Selection and temporal correlation.} Simply retaining less data does not explain the gain. Random retention yields 3.770 bottles/scene, slightly below vanilla BC; WARP-BC places 0.764 more bottles/scene at the same budget (95\% CI [0.60, 0.93]). The time-warp structure also matters: replacing correlated AR(1) log-speeds with IID draws lowers throughput from 290 to 269 bottles/hr (7\%) without changing retention. These comparisons support both selecting productive chunks and learning the selection signal with temporally correlated warps. \ifcorlcameraready\else
An additional training seed preserves the ordering of WARP-BC and DemInf (287 vs.\ 261 bottles/hr; 4.492 vs.\ 4.172 bottles/scene). ReWiND reaches 3.781 bottles/scene, close to random retention. These comparisons support the structured relative-progress signal without establishing a universal advantage across tasks.

\fi

\textbf{Open reproduction.} The project page provides data, checkpoints, n=512 traces, and pinned code for end-to-end WARP reproduction, and documents baseline-specific reproduction limits.

\subsection{Direct Validation of the Progress Signal}
\label{sec:signal_validation}
To test what the score recognizes, we compare its predictions with annotations collected before this work by independent annotators across 452 expert T-shirt episodes. Predicted velocity is lower during mistakes than in duration-matched productive windows from the same episode in 23/24 cases, including 12/13 outside $\mathcal{D}_{RM}$. Frame-level mistake-discrimination AUROC is 0.83, versus 0.61 for proprioceptive speed and 0.68 for garment-area change. The small annotated sample supports discrimination between productive and mistaken behavior; it does not establish that curation preserves useful recovery segments.

\section{Limitations and Future Work}
\label{sec:limitations}
While \bcmethod{} attenuates the impact of suboptimal data, the resulting policy remains restricted to the behaviors present in the training set. Future work will investigate how DAgger and post-training, e.g. offline and online RL, and iterative self-improvement via reward-aligned learning, can generalize performance beyond these initial offline demonstrations. Furthermore, our experiments span two real manipulation tasks and one simulated task on a single bimanual embodiment, and several directions remain open. First, our negative-progress supervision comes entirely from reversed playback, an approximation we adopt from ReWiND~\cite{zhang2025rewind}, which notes that such reversals can be physically implausible. This introduces a distribution gap: a real-world regression unfolds forward in time under normal causal dynamics, whereas temporal-warping and reversals present a different spatiotemporal distribution. We demonstrate that this gap frequently exhibits sufficient overlap in practice to extract viable signal for reward learning. Still, this alignment is task-dependent, and we encourage practitioners to validate whether this synthetic signal yields useful representations on their own tasks.




\clearpage
\acknowledgments{This research was performed at the AUTOLAB at UC Berkeley in affiliation with the Berkeley AI Research (BAIR) Lab. XDOF provided the teleoperated demonstration data, training compute, and evaluation hardware infrastructure used for this project. In their academic roles at UC Berkeley, Justin Yu, Andrew Goldberg, Kavish Kondap, Karim El-Refai, Ethan Ransing, and Ken Goldberg are supported in part by donations from Toyota Research Institute, Autodesk, Meta, Google, Siemens, and Bosch, and by equipment grants from NVIDIA, PhotoNeo, the NSF AI4OPT Center, and Intuitive Surgical. Justin Yu is supported by the National Science Foundation Graduate Research Fellowship Program under Grant No.\ DGE 2146752. Any opinions, findings, and conclusions or recommendations expressed in this material are those of the authors and do not necessarily reflect the views of the National Science Foundation.}


\bibliography{example}  

\newpage
\appendix

\section{Dataset Statistics}
\label{app:data}

\begin{figure}[ht]
\centering
\includegraphics[width=0.95\linewidth]{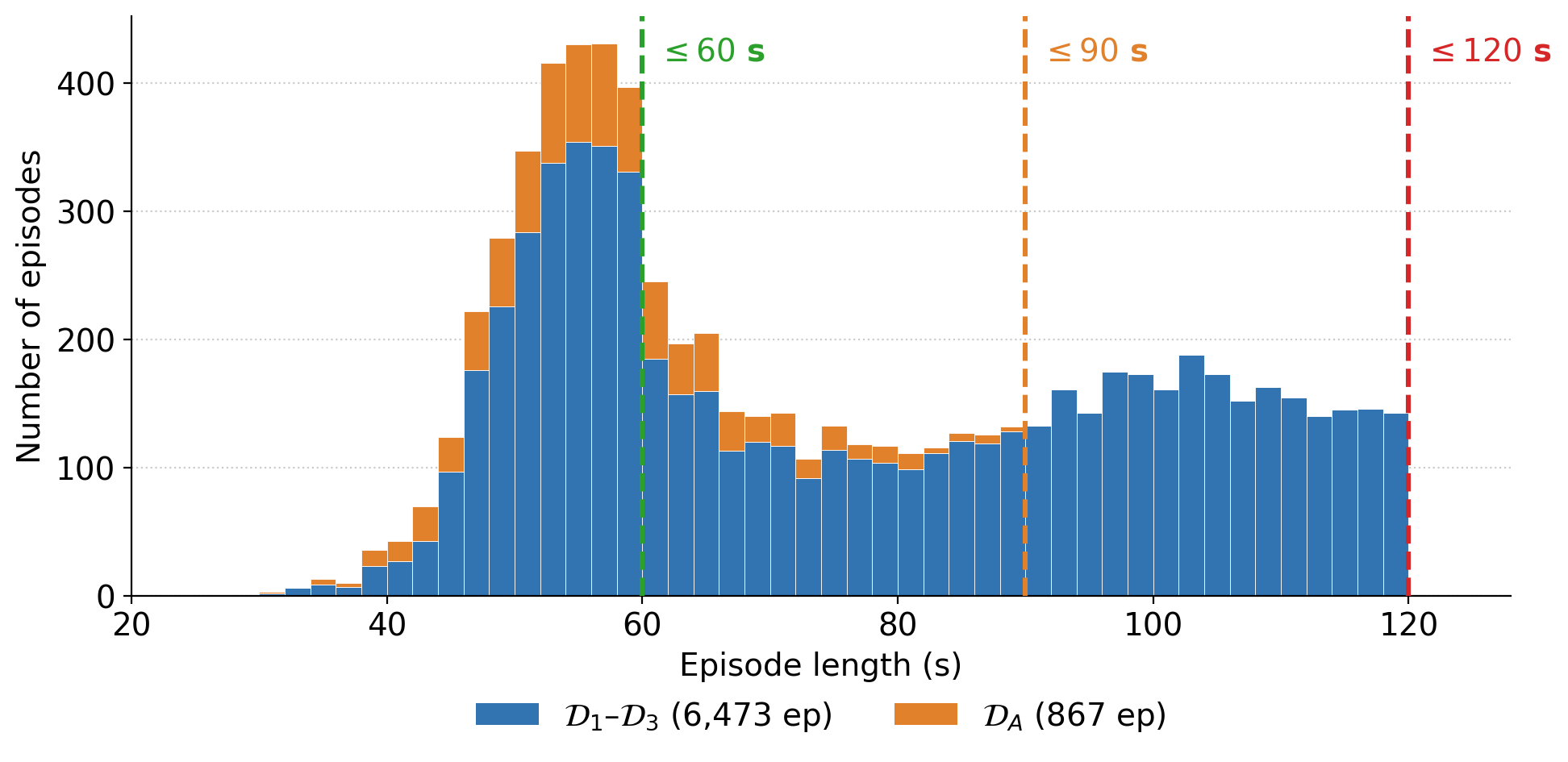}
\caption{\textbf{Episode-length distribution of $\mathcal{D}_1$--$\mathcal{D}_3$ (blue) with
  the SARM-annotated supplement $\mathcal{D}_A$ overlaid (orange).} The base distribution exhibits a dominant mode near $50$--$60$\,s with a broader tail beyond $\sim\!85$\,s containing episodes with more hesitations, fumbles,
  and recoveries. Dashed vertical lines mark the three length filters used in
  Section~\ref{sec:result}; unioning $\mathcal{D}_A$ with $\mathcal{D}_1$ and
  $\mathcal{D}_2$ yields the matched datasets $\mathcal{D}_4 = \mathcal{D}_1 \cup \mathcal{D}_A$ and $\mathcal{D}_5 = \mathcal{D}_2 \cup \mathcal{D}_A$ used in the SARM comparison. The \rewardmethod{} reference subset
  (Table~\ref{tab:appendix_corpus}) lies entirely to the left of the green $\le 60$\,s line.}
\label{fig:episode_length_histogram}
\end{figure}

Table~\ref{tab:appendix_corpus} reports per-tier statistics for the three policy training datasets used in Section~\ref{sec:result}, as well as the fixed reference subset on
which \rewardmethod{} is trained. All tiers are length-filtered subsets of a single underlying dataset of human-teleoperated bimanual T-shirt-folding demonstrations.
Figure~\ref{fig:episode_length_histogram} plots the corresponding episode-length distribution.

\begin{table}[ht]
  \centering
  \small
  \setlength{\tabcolsep}{10pt}
  \renewcommand{\arraystretch}{1.25}
  \begin{tabular}{l l r r}
  \toprule
  \textbf{Dataset / Tier} & \textbf{Description} & \textbf{Episodes} & \textbf{Total hours} \\
  \midrule
  $\mathcal{D}_1$           & Policy training, length-filter $\le 60$\,s   & $2{,}427$ & $36.1$ \\
  $\mathcal{D}_2$           & Policy training, length-filter $\le 90$\,s   & $4{,}124$ & $71.3$ \\
  $\mathcal{D}_3$           & Policy training, length-filter $\le 120$\,s  & $6{,}473$ & $139.7$ \\
  $\mathcal{D}_A$           & Annotated dataset, SARM train data & $867$  & 13.9 \\
  \midrule
  $\mathcal{D}_{RM}$ & WARP-RM train data, $\le 59.8$\,s       & $1{,}950$ & $28.7$ \\
  \bottomrule
  \end{tabular}
  \vspace{5pt}
  \caption{\textbf{Per-tier dataset statistics.} Tiers $\mathcal{D}_1$--$\mathcal{D}_3$ admit progressively more suboptimal demonstrations under a length-based proxy.
  Tiers $\mathcal{D}_4$--$\mathcal{D}_5$ are augmented with $\mathcal{D}_A$ used for the SARM RA-BC comparison (Table~\ref{tab:sarm_compare}). The \rewardmethod{}
  reference subset is held fixed across all $\mathcal{D}_i$ policy-tier experiments and consists of the shortest, fastest demonstrations in the dataset, providing a clean training signal for the canonical $\hat v = 1$ progress unit (Sec.~\ref{sec:method_progress}). All footage at $30$\,Hz.}
  \label{tab:appendix_corpus}
\end{table}

\section{Bottle-in-Bin Dataset Statistics}
\label{app:bottle_data}

\begin{figure}[ht]
\centering
\includegraphics[width=0.95\linewidth]{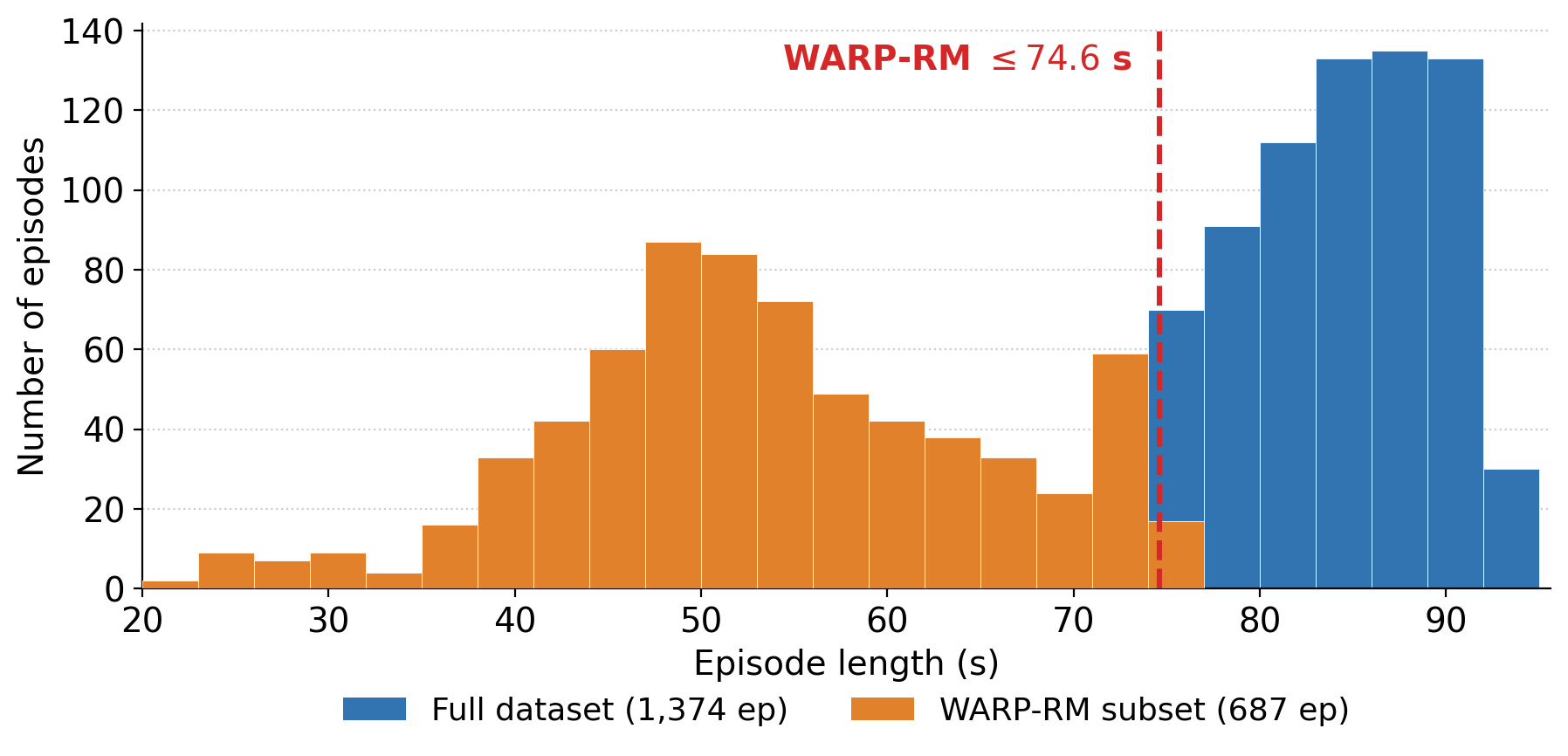}
\caption{\textbf{Episode-length distribution of the bottle-in-bin dataset.} \progressmodel{} is trained on the shortest
  demonstrations (orange, $\le 74.6$\,s); the dashed line marks the cutoff. As in the T-shirt setting
  (Fig.~\ref{fig:episode_length_histogram}), \progressmodel{} is trained only on the shortest, fastest demonstrations.}
\label{fig:bottle_length_histogram}
\end{figure}

Table~\ref{tab:bottle_corpus} and Figure~\ref{fig:bottle_length_histogram} report dataset statistics for the
bottle-in-bin placement task (Section~\ref{sec:result}, Table~\ref{tab:bottle_results}). Both the vanilla BC and
\bcmethod{} policies are trained on the dataset of human-teleoperated ``put the plastic bottles in the bin'' demonstrations.
As with the T-shirt dataset, \progressmodel{} is trained once on a fixed reference subset of the shortest demonstrations
($\le 74.6$\,s), providing a clean signal for the canonical execution pace ($\hat v = 1$).
\textbf{All \rewardmethod{} and \progressmodel{} hyperparameters are kept identical to the T-shirt setup}
(Tables~\ref{tab:warp_hparams}--\ref{tab:warp_rm_hparams}); only the training dataset differs.

\begin{table}[ht]
  \centering
  \small
  \setlength{\tabcolsep}{10pt}
  \renewcommand{\arraystretch}{1.25}
  \begin{tabular}{l l r r}
  \toprule
  \textbf{Dataset} & \textbf{Description} & \textbf{Episodes} & \textbf{Total hours} \\
  \midrule
  Full dataset    & Human-teleoperated demonstrations                           & $1{,}374$ & $26.4$ \\
  \midrule
  WARP-RM subset  & \rewardmethod{}-RM train data, shortest demos $\le 74.6$\,s     & $687$     & $10.2$ \\
  \bottomrule
  \end{tabular}
  \vspace{5pt}
  \caption{\textbf{Bottle-in-bin dataset statistics.} The \rewardmethod{} reference subset consists of the shortest
  demonstrations ($\le 74.6$\,s), mirroring the T-shirt protocol (Table~\ref{tab:appendix_corpus}) and providing a clean
  training signal for the canonical $\hat v = 1$ progress unit (Sec.~\ref{sec:method_progress}). All footage at $30$\,Hz.}
  \label{tab:bottle_corpus}
\end{table}

\section{Simulation Bottle-in-Bin Dataset Statistics}
\label{app:sim_bottle_data}

\begin{figure}[ht]
\centering
\includegraphics[width=0.95\linewidth]{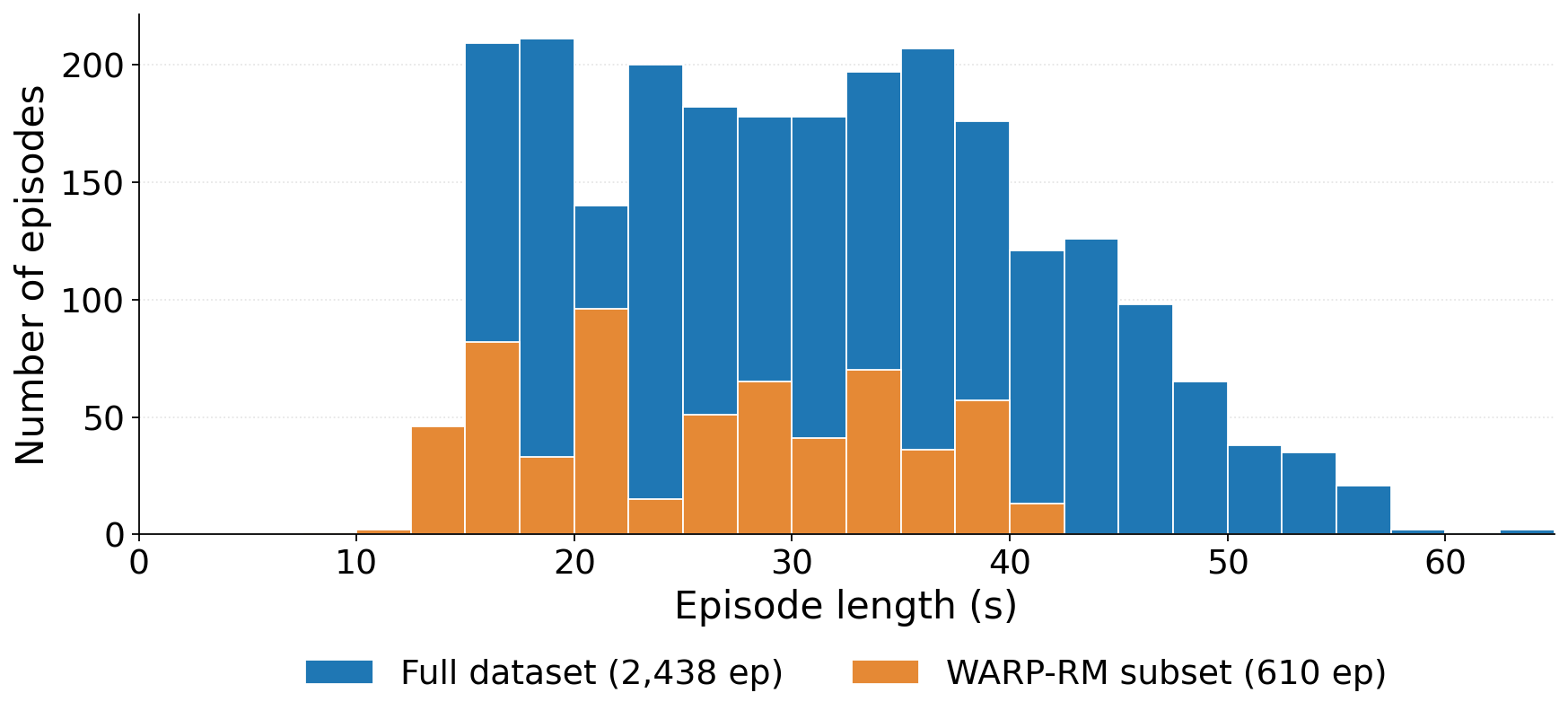}
\caption{\textbf{Episode-length distribution of the simulation bottle-in-bin dataset.}
  \progressmodel{} is trained on the shortest demonstrations (stratified per bottle count).}
\label{fig:sim_bottle_length_histogram}
\end{figure}

Table~\ref{tab:sim_bottle_corpus} and Figure~\ref{fig:sim_bottle_length_histogram} report dataset
statistics for the simulated bottle-in-bin placement task (Section~\ref{sec:sim_bottles},
Table~\ref{tab:sim_bottle_results}). All policies in the comparison are trained on
the same corpus of simulated ``put the plastic bottles in the bin'' demonstrations, collected in
the MuJoCo YAM environment and rendered at $30$\,Hz. Dataset scenes contain between two and six
bottles (near-uniformly distributed), but evaluation scenes use six bottles consistently. As in the real-world setting, \progressmodel{} is
trained once on a fixed reference subset of the shortest demonstrations --- here the shortest
$25\%$, stratified per bottle count so that faster small-count episodes do not crowd out
six-bottle episodes ($610$ episodes, all $\le 40$\,s) --- providing a clean signal for the
canonical execution pace ($\hat v = 1$). The simulation uses $S=0.5$\,s (15 source frames at 30\,Hz), so a 32-token inference window spans 15.5\,s. The absolute sampler path budget remains fixed rather than being rescaled with $S$; App.~\ref{app:sim_protocol} specifies the differences from the real-world recipe.

\begin{table}[ht]
  \centering
  \small
  \setlength{\tabcolsep}{10pt}
  \renewcommand{\arraystretch}{1.25}
  \begin{tabular}{l l r r}
  \toprule
  \textbf{Dataset} & \textbf{Description} & \textbf{Episodes} & \textbf{Total hours} \\
  \midrule
  Full dataset    & Simulated bottle-in-bin demonstrations                       & $2{,}438$ & $17.5$ \\
  \midrule
  WARP-RM subset  & \rewardmethod{}-RM train data & $610$     & $3.4$ \\
  \bottomrule
  \end{tabular}
  \vspace{5pt}
  \caption{\textbf{Simulation bottle-in-bin dataset statistics.} The \rewardmethod{} reference
  subset consists of the shortest demonstrations, mirroring the
  real-world protocol (Table~\ref{tab:bottle_corpus}) and providing a clean training signal for
  the canonical $\hat v = 1$ progress unit (Sec.~\ref{sec:method_progress}). All footage at
  $30$\,Hz.}
  \label{tab:sim_bottle_corpus}
\end{table}

\section{Simulation Training and Evaluation Protocol}
\label{app:sim_protocol}
\textbf{Reward model.} The WARP reward model uses the shortest 25\% of demonstrations stratified by bottle count, the top camera, feature stride one, $N=32$, and $S=0.5$\,s. Training runs for 20,000 steps; the released head is the best-composite checkpoint at step 14,400. The AR(1) path-budget center and half-range remain 1.5\,s and 1.0\,s per gap, respectively: $\ell\in[465,2325]$ source frames before episode-length clipping. The categorical support is fixed at $[-3,3]$. The IID ablation changes log-speed correlation while retaining the same path budget and reversal process.

\textbf{Policy training.} All simulation policies use $\pi_0$ initialized from the public base checkpoint, batch size 32, 30,000 training steps, cosine learning-rate decay, and action horizon $H=30$. Retention thresholds are calibrated per method to 31.5\%; vanilla uses 100\%. WARP-BC and its IID variant use binary filtering, with unit loss weight for every retained chunk. Thus the simulation comparison evaluates WARP-based selection, without continuous reweighting within the retained subset. The candidate-anchor denominator includes every source frame, with episode-tail windows padded using the final frame. Holding retention and training steps fixed controls effective data reuse across curation methods. In contrast, vanilla revisits a larger retained dataset.

\textbf{Rollouts.} Each arm uses the same 512 scene seeds. Every predicted 30-action chunk is executed in full, with replanning on chunk exhaustion. The physics timestep is 0.002\,s with control decimation 17, giving 0.034\,s per action and 61.2\,s for 1,800 actions. The scorer uses recorded physics states to identify qualifying bottle placements. Reported confidence intervals compare methods at the paired-scene level. The additional-seed comparison in Sec.~\ref{sec:sim_bottles} is reported separately from the primary table.

\textbf{Public artifacts.} The reproduction guide at \url{https://github.com/uynitsuj/WARP-RM} links the dataset, checkpoints, and trace releases. The policy-training and batched-evaluation code is pinned to \texttt{uynitsuj/openpi@9c8e7b75} and \texttt{uynitsuj/abc-rabc@59db543d}. The historical n=128 trace self-test is a deterministic scorer check; it must not be substituted for the n=512 comparison. The guide documents which baseline curation pipelines can be regenerated and which have verification-only support.

\section{Implementation Details}
  \label{app:impl}

  \subsection{\rewardmethod{} Hyperparameters (Real Experiments)}

  \begin{table}[H]
  \centering
  \small
  \setlength{\tabcolsep}{6pt}
  \renewcommand{\arraystretch}{1.15}
  \begin{tabular}{lll}
  \toprule
  \textbf{Group} & \textbf{Symbol / name} & \textbf{Value} \\
  \midrule
  \multirow{6}{*}{Sampler (Eq.~\ref{eq:ar1}--\ref{eq:rescale})}
    & Window length $N$                & $32$ frames \\
    & Canonical inter-token stride $S$ & $1.5$\,s (45 frames @ $30$\,Hz) \\
    & AR(1) autocorrelation $\alpha$   & $0.5$ \\
    & Marginal log-velocity std $\sigma_\infty$ & $\ln 2$ \\
    & Path-length mean $\bar{s}$       & $\mathrm{Uniform}([0.5,\; 2.5])$ seconds per gap\\
    & Reversal rate $\lambda_{\text{rev}}$ & $1$ (Poisson) \\
    & Global time-reverse prob.\ $p_{\text{flip}}$ & $0.5$ \\
  \bottomrule
  \end{tabular}
  \vspace{5pt}
  \caption{\textbf{\rewardmethod{} Sampler Configuration.}}
  \label{tab:warp_hparams}
  \end{table}

  \subsection{\progressmodel{} Hyperparameters}

  \begin{table}[H]
  \centering
  \small
  \setlength{\tabcolsep}{6pt}
  \renewcommand{\arraystretch}{1.15}
  \begin{tabular}{lll}
  \toprule
  \textbf{Group} & \textbf{Symbol / name} & \textbf{Value} \\
  \midrule
  \multirow{3}{*}{Targets (Eq.~\ref{eq:rel_label})}
    & Framerate $f$  & 30 Hz \\
    & Normalization $C_{\text{norm}}$  & 
    $(N{-}1)\cdot S \cdot f = 1395$ source-frames \\
    & Output bins                       & $30$, centers linearly spaced in $[-3, 3]$ \\
    & Target encoding                   & two-hot \\
  \midrule
  \multirow{6}{*}{Architecture}
    & Visual backbone $\phi$            & frozen DINOv3 ViT-B/16, $768$-dim \\
    & Input resolution                  & $224 \times 224$ \\
    & Temporal-diff projection          & $\mathbb{R}^{1536} \to \mathbb{R}^{768}$ (no bias) \\
    & Encoder layers / heads / model dim & $12$ / $8$ / $768$ \\
    & Attention                         & bidirectional self-attention \\
    & Dropout                           & $0.15$ \\
    & Positional embedding              & fixed sinusoidal \\
  \midrule
  \multirow{6}{*}{Optimization}
    & Optimizer                         & AdamW ($\beta_1{=}0.9$, $\beta_2{=}0.999$) \\
    & Peak learning rate                & $4 \times 10^{-4}$ (linear-scaled from $10^{-4}$ at bs $= 256$) \\
    & Weight decay                      & $10^{-3}$ \\
    & Batch size                        & $1024$ windows \\
    & Total steps                       & $15\,000$ \\
    & Warmup / schedule                 & $1000$ linear warmup, cosine anneal to $0$ \\
    & Gradient clip                     & $1.0$ \\
  \bottomrule
  \end{tabular}
  \vspace{5pt}
  \caption{\textbf{\progressmodel{} Training Configuration.} All values are held fixed across the three policy tiers $\mathcal{D}_1, \mathcal{D}_2, \mathcal{D}_3$;
  \progressmodel{} is trained once on $\mathcal{D}_{RM}$ and reused as a frozen scorer.}
  \label{tab:warp_rm_hparams}
  \end{table}
  
  \subsection{\bcmethod{} Policy Training}

  We instantiate the behavior cloning policy with the $\pi_0$ flow-matching backbone~\cite{black2026pi0visionlanguageactionflowmodel}. Action chunks span $H = 30$
  source-frames ($1.0$\,s); per-chunk weights $w$ are precomputed once per episode using Eq.~\ref{eq:warpbc_weight} with threshold $\tau = 1.0$, and chunks with $w = 0$ are
  removed from the dataset at construction time so they consume no compute. All other policy-side hyperparameters (optimizer, schedule, batch size, image augmentation)
  follow the public $\pi_0$ recipe unchanged.

  \subsection{Dense Inference for Reweighting}

  To produce a per-frame velocity $\hat{v}_t$ over a target policy training dataset, we run sliding-window inference at stride-$1$ source-frame offsets, each window using the
  canonical $S = 1.5$s inter-token spacing. The $(N{-}1) \cdot S \cdot f$ overlapping per-window velocities covering source-frame $t$ are averaged to give $\hat{v}_t$.
  This is done once per episode before policy training begins; the resulting $\hat{v}_t$ trace is cached alongside the dataset.

\newpage
\section{Evaluation Protocol Details}
\label{app:eval_protocol}
Trials in which the bin is knocked over (bottle task) or grasped instead of a garment within the first 10 seconds (folding task) are reset and repeated rather than scored, as the demonstrations contain no bin-recovery behavior. The same rule applies to all methods.

\newpage
\section{Additional Experimental Result Statistics}
\label{app:exp_statistics}

\begin{figure}[H]
\centering
\includegraphics[width=\textwidth]{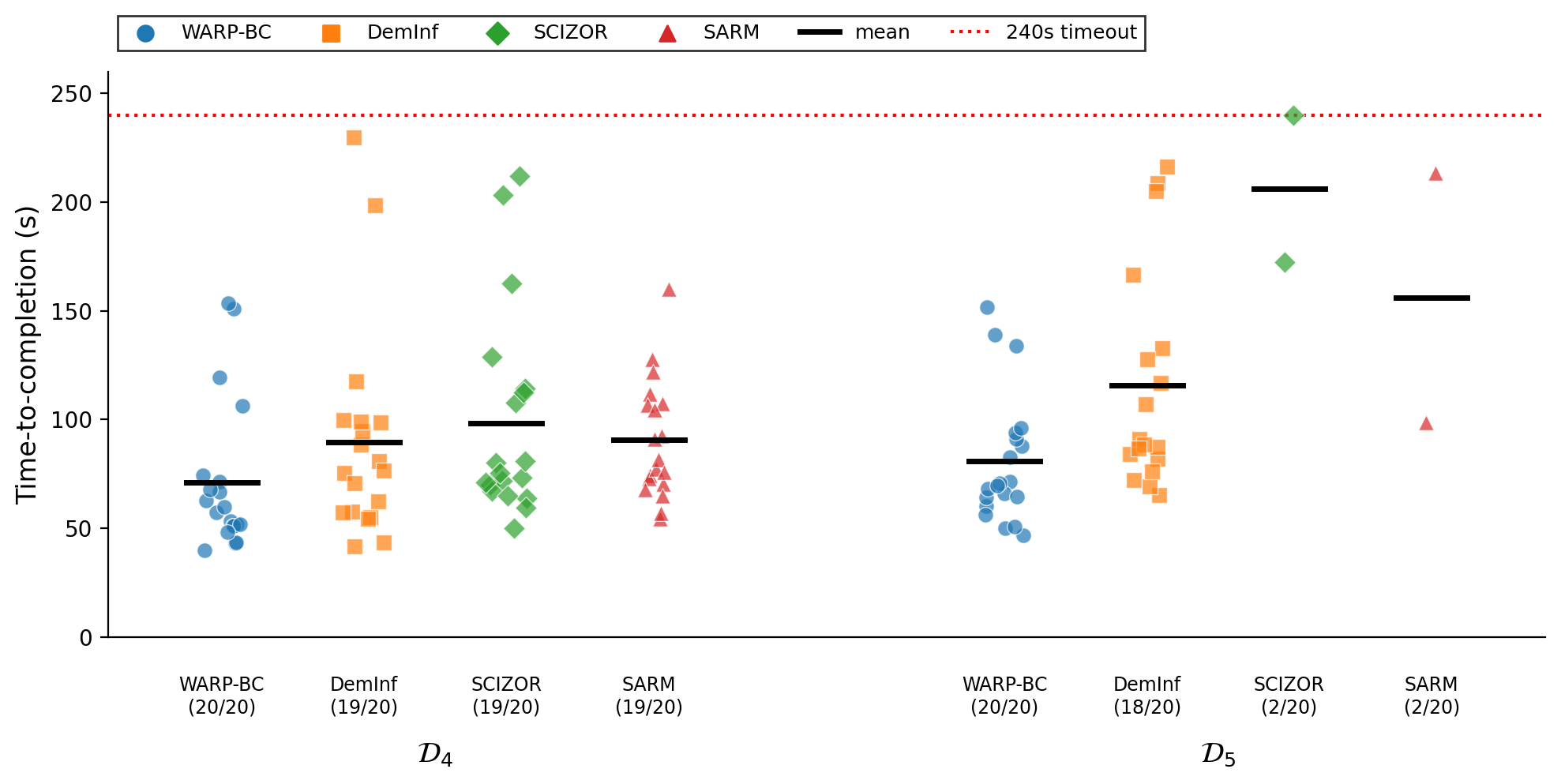}
\caption{\textbf{Time-to-completion distribution for success across baseline comparisons.} Evaluated on $\mathcal{D}_4 = \mathcal{D}_1 \cup \mathcal{D}_A$ and $\mathcal{D}_5 = \mathcal{D}_2 \cup \mathcal{D}_A$. Policy rollouts which exceed 240 seconds are considered failures and are not shown. SCIZOR~\cite{zhang2025scizor} successfully folds a T-shirt right before the 240 second timeout boundary on $\mathcal{D}_5$}.
\label{fig:ttc_baseline_distribution}
\end{figure}

\newpage
\section{WARP-RM Qualitative Results}
\label{app:qualitative}
\begin{figure}[H]
\centering
\includegraphics[width=\textwidth]{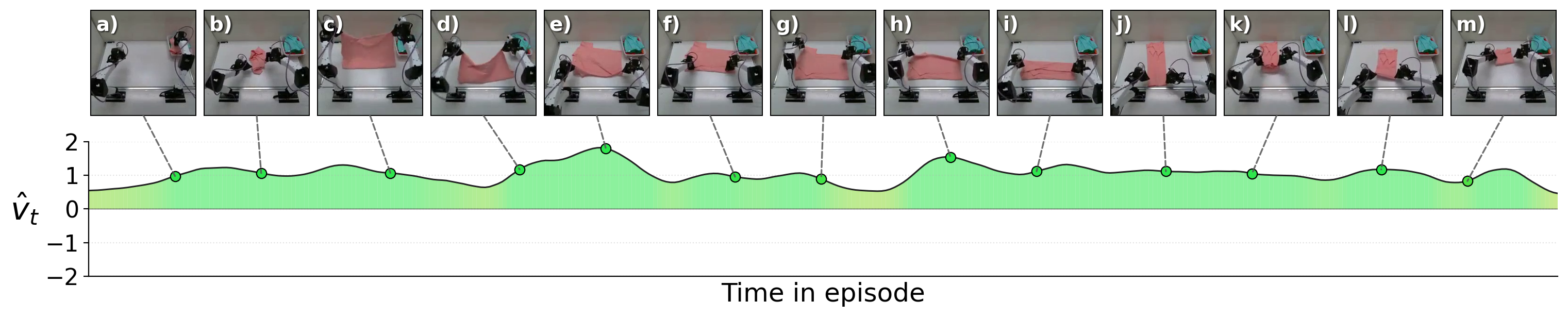}
\caption{\textbf{WARP-RM output on a near-unit average progress-velocity T-shirt-folding demonstration.} Predicted magnitude varies around 1.0 for most of the demonstration. (34 second demonstration).}
\end{figure}

\begin{figure}[H]
\centering
\includegraphics[width=\textwidth]{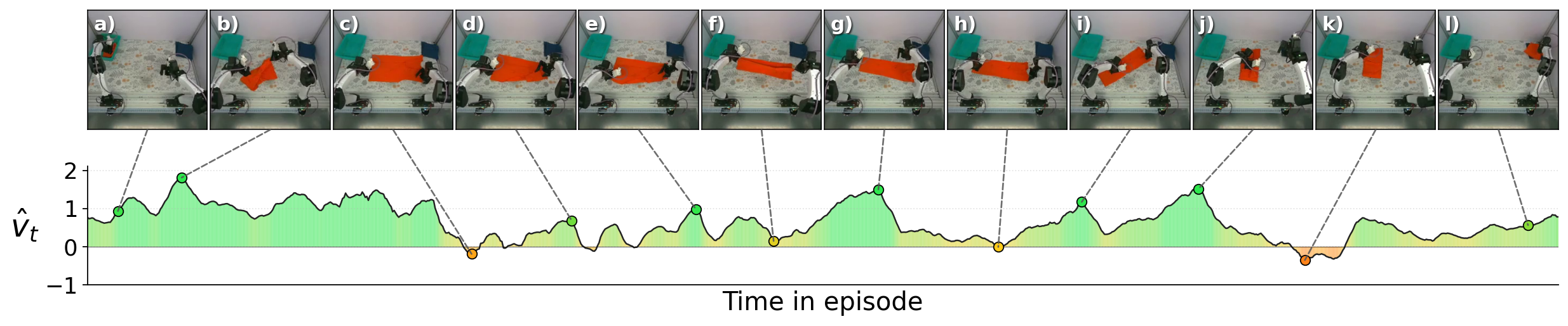}
\caption{\textbf{WARP-RM output on a T-shirt-folding demonstration with fluctuating progress-velocity.} (97 second demonstration).}
\end{figure}

\begin{figure}[H]
\centering
\includegraphics[width=\textwidth]{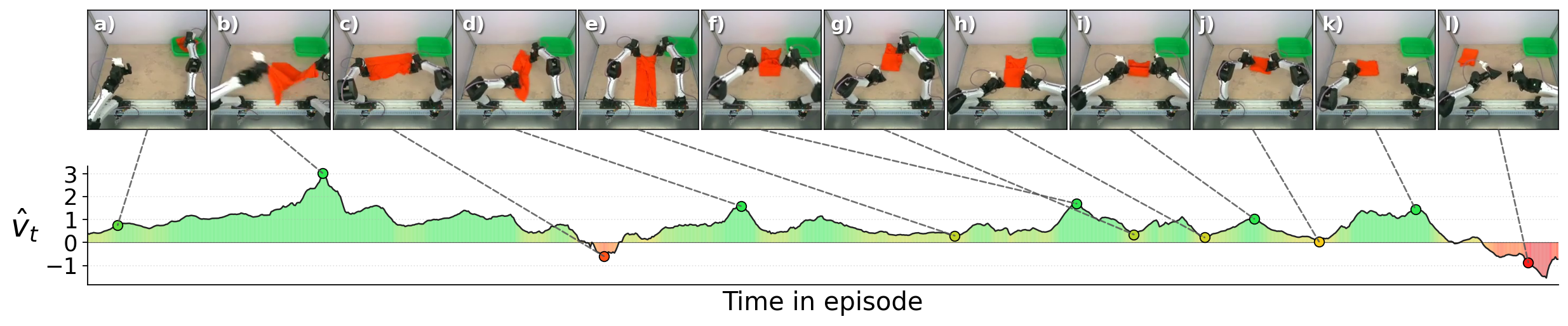}
\caption{\textbf{WARP-RM output on a T-shirt-folding demonstration with fluctuating progress-velocity.} (98 second demonstration).}
\end{figure}

\begin{figure}[H]
\centering
\includegraphics[width=\textwidth]{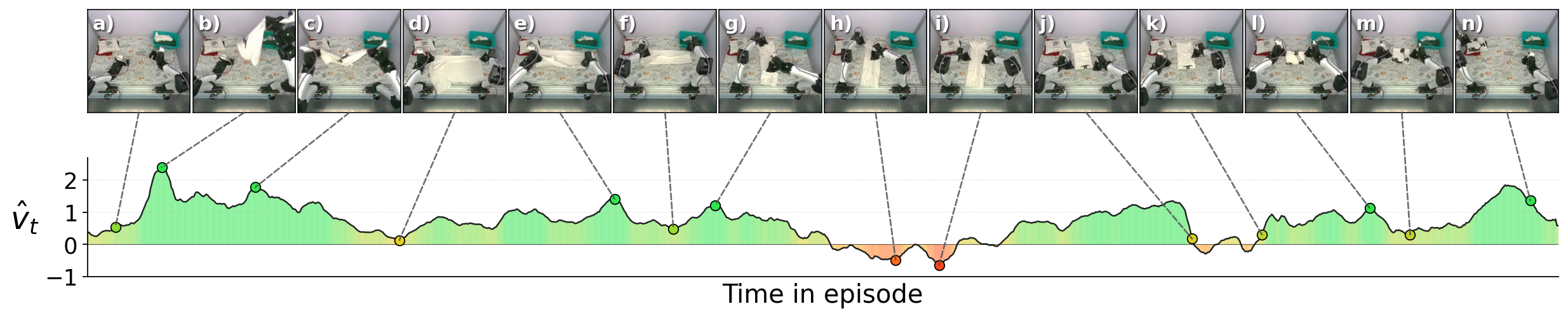}
\caption{\textbf{WARP-RM output on a T-shirt-folding demonstration  with fluctuating progress-velocity.} (105 second demonstration).}
\end{figure}

\newpage
\section{Training Dataset Visual Diversity}
\begin{figure}[H]
\centering
\includegraphics[width=\textwidth]{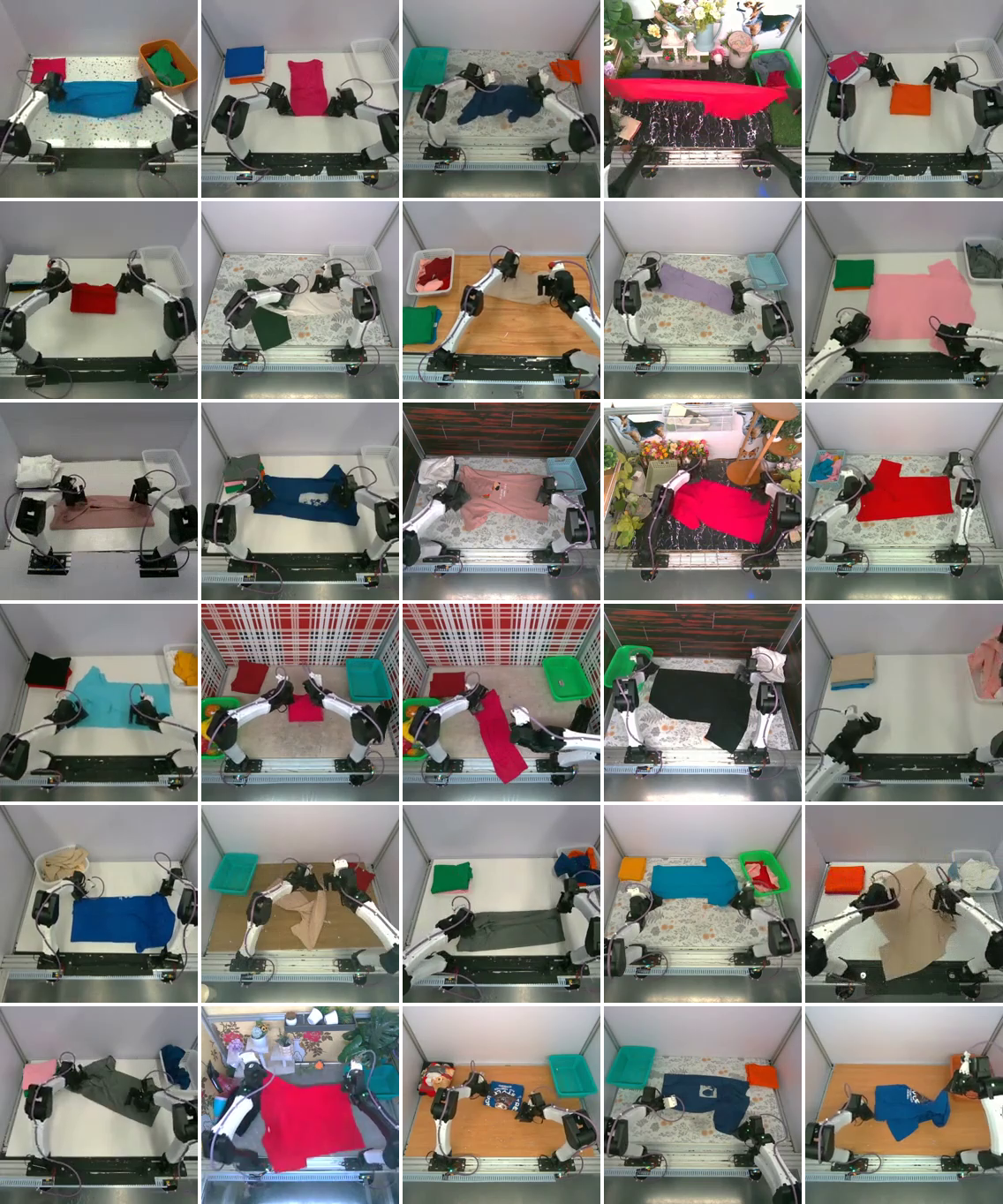}
\caption{\textbf{Randomly sampled frames from the T-shirt-folding dataset ($\mathcal{D}_3$)}, demonstrating a representative sample of the visual diversity present in the training data, including varied garment colors, workspace surfaces, and arm configurations.}
\end{figure}

\begin{figure}[H]
\centering
\includegraphics[width=\textwidth]{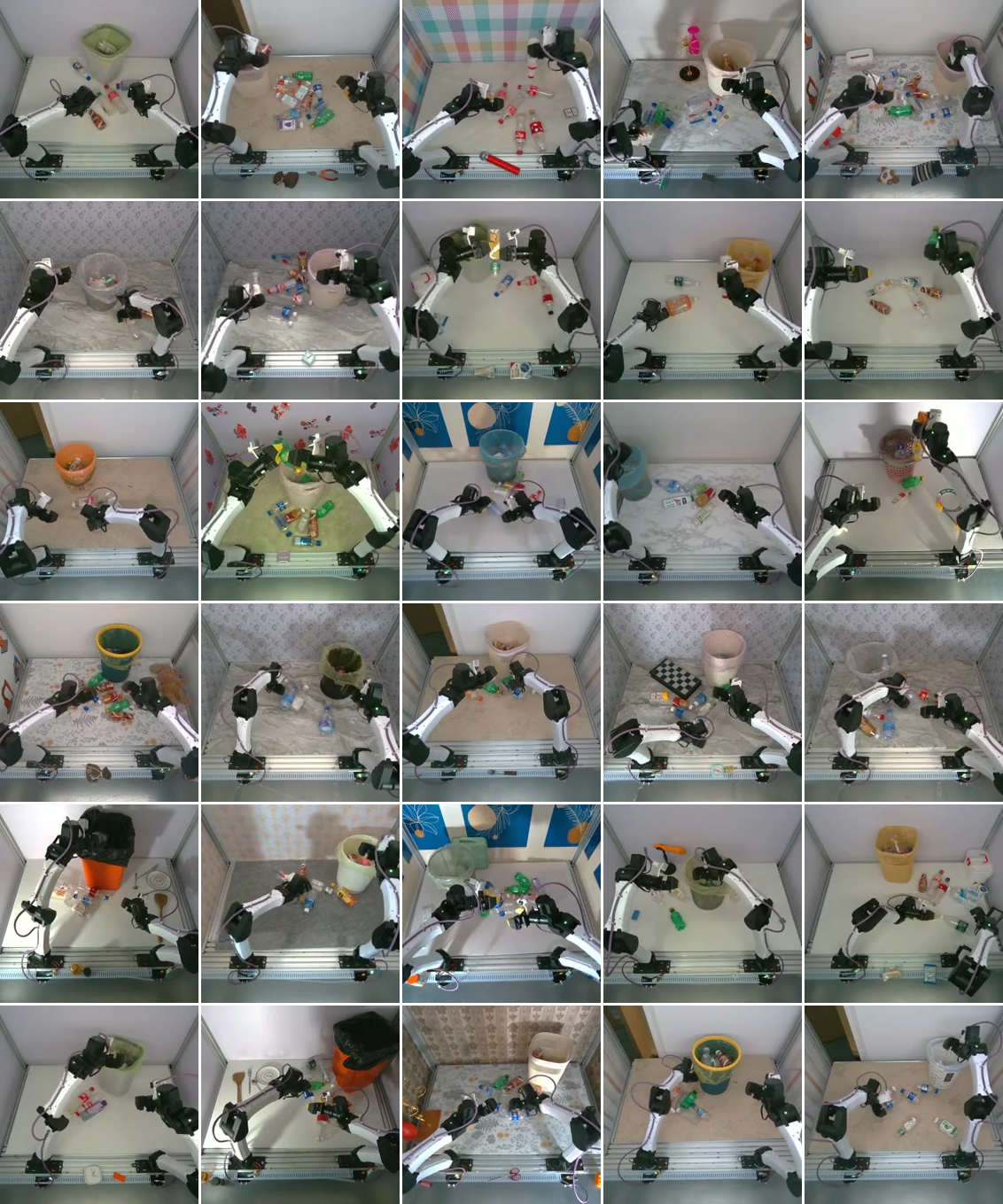}
\caption{\textbf{Randomly sampled frames from the bottle-in-bin dataset}, drawn from demonstrations across distinct collection sessions. The data spans varied bin types and placements, bottle colors and counts, and workspace surfaces.}
\end{figure}

\end{document}